\documentclass{article}

\PassOptionsToPackage{numbers,square}{natbib}
\usepackage[preprint]{stylee}
\usepackage{tikz}
\usepackage{pgfplots}
\pgfplotsset{compat=1.17}
\usetikzlibrary{patterns,fillbetween,arrows.meta}
\usepgfplotslibrary{groupplots}
\usepackage{cancel}
\usepackage{array, tabularx, booktabs}
\usepackage{placeins}
\usepackage{float}
\usepackage[normalem]{ulem}
\usepackage{coffeestains}
\usepackage{graphicx}
\usepackage{subcaption} 
\usepackage{bm}
\usepackage{amsmath,amssymb,amsthm}     
\usepackage{tcolorbox}
\usepackage{enumitem} 

\newcounter{boxnum}
\newenvironment{discussion_box}[1][Key Takeaways and Recommendations]{%
  \refstepcounter{boxnum}
  \begin{tcolorbox}[title={Box \theboxnum: #1},%
    colback=gray!5,
    colframe=black,
    fonttitle=\bfseries,
    boxrule=0.8pt,
    arc=2mm,
    left=6pt,
    right=6pt,
    top=6pt,
    bottom=6pt]%
}{%
  \end{tcolorbox}%
}

\usepackage[utf8]{inputenc} 
\usepackage[T1]{fontenc}    
\usepackage{hyperref}       
\usepackage{url}            
\usepackage{booktabs}       
\usepackage{amsfonts}       
\usepackage{nicefrac}       
\usepackage{microtype}      
\usepackage{xcolor}         
\usepackage{siunitx}
\usepackage{amsmath}

\title{Collapsibility of Performance Metrics\\ in Clinical Predictive AI\\}

\author{%
  João Matos \thanks{\texttt{joao.matos@ndorms.ox.ac.uk}, Centre for Statistics in Medicine, University of Oxford} \\
  University of Oxford\\
  \And
    Ben Van Calster \\
    KU Leuven\\
  \And
    Richard D. Riley \\
    University of Birmingham\\\\
  \AND
  Paula Dhiman \\
    University of Oxford\\
  \And
  Gary S. Collins \\
University of Birmingham\\
}

\begin{document}
\maketitle

\medskip
\medskip

\begin{abstract}
\textbf{Background:} Population level assessments of predictive artificial intelligence (AI) can conceal performance disparities across subgroups. Fairness evaluations commonly rely on performance analyses across subgroups. However, some performance metrics are \textit{non-collapsible}, meaning that the overall population performance value does not equal the weighted average of subgroup specific values. This can mislead fairness evaluation.
\vspace{0.5em}

\textbf{Objective:} To examine the collapsibility properties of commonly reported performance metrics in predictive AI, with a focus on the area under the receiver operating characteristic curve (AUC, also known as c-statistic).

\vspace{0.5em}

\textbf{Methods:} We investigate the collapsibility of fifteen performance metrics, either by expressing each metric as a linear combination of its stratum specific values or, where non-collapsible, by providing a counterexample inspired by Simpson's paradox as a formal disproof. We focus particularly on the AUC, a commonly reported metric of model discrimination, and formally characterise its expression under mixtures of subpopulations. 

\vspace{0.5em}

\textbf{Results:} Five performance metrics (AUC, calibration intercept, calibration slope, expected calibration error, and Nagelkerke $R^2$) are shown to be non-collapsible, and ten (O:E ratio, logloss, Brier score, accuracy, F1-score, true positive rate, true negative rate, positive predictive value, negative predictive value, and net benefit) are shown to be collapsible. The AUC is shown to be non-collapsible because it decomposes into within- and cross-group AUC terms when subpopulations coexist, such that its overall value may fall outside the range of subgroup specific AUCs. 

\vspace{0.5em}

\textbf{Conclusions:} Non-collapsibility of performance metrics has important consequences for reporting, model appraisal, and fairness evaluation. It can generate spurious differences between subgroup and overall performance, which may mislead fairness evaluations. Explicitly acknowledging and reporting the collapsibility properties of performance metrics improves both the interpretability and transparency of fairness assessments.
\end{abstract}

\newpage

\begin{discussion_box}[Glossary of key terms used throughout the manuscript]\label{box:second}
\small

\textbf{AUC:} area under the receiver operating characteristic curve, also known as c-statistic \cite{janssens2020reflection}.

\medskip \textbf{Calibration:} the extent of the agreement between the probability estimates and the observed event proportions \cite{van2016calibration}.

\medskip \textbf{Collapsibility:} a metric is said to be collapsible over a stratifying variable $g$ (e.g., sex) if its marginal value equals the weighted average of the stratum specific values (e.g., male, female), with weights depending on the marginal distribution of $g$ \cite{huitfeldt2019collapsibility}.

\medskip \textbf{Cross-group discrimination:} $\mathrm{xAUC_{i,j}}$, probability that an individual with the event from subgroup $i$ is ranked higher than an individual without the event from subgroup $j$ \cite{Kallus2019-yl}.

\medskip \textbf{Discrimination:} extent to which the model gives higher probability estimates for individuals with the event compared to those without the event, quantified by the area under the receiver operating characteristic curve (AUC) \cite{Van-Calster2024-jw}. 

\medskip \textbf{Fairness evaluation:} assessing the nature and magnitude of any potential differential model performance across individuals or subgroups defined by a sensitive attribute. It can be conducted in different ways, including disaggregated performance evaluation; stratification of visual tools such as calibration plots; computation of fairness metrics.

\medskip \textbf{Fairness metric:} a metric that quantifies the extent to which a model’s output does not discriminate (in the societal sense, based on a given notion of fairness) against individuals or subgroups defined by a sensitive attribute \cite{matos2025critical}.

\medskip \textbf{Jensen's Gap:} difference between the two sides of Jensen's inequality, i.e., the deviation of the overall metric from the linear interpolation of subgroup values \cite{denny2017fallacy}.

\medskip \textbf{Outcome prevalence ($\pi$):} proportion of individuals experiencing an outcome of interest.

\medskip \textbf{Simpson's Paradox:} phenomenon in which a trend observed within multiple subgroups reverses or disappears when subgroups are combined \cite{blyth1972simpson}.

\medskip \textbf{Subgroup proportion ($p^g$):} proportion of individuals belonging to subgroup $g$ within the population.

\medskip \textbf{Within-group discrimination ($\mathrm{AUC_g}$):} probability that an individual with the event from subgroup $g$ is ranked higher than an individual without the event from subgroup $g$ \cite{Kallus2019-yl}.

\end{discussion_box}

\newpage

\section{Introduction}

Evaluating the performance of a clinical prediction model at the population level can conceal important differences in model performance across subgroups. To address this, fairness assessments often report performance within relevant subgroups defined by a sensitive attribute, such as sex or ethnicity, or compute explicit fairness metrics \cite{matos2025critical,van2026navigating}. Even so, interpreting subgroup specific results (e.g, male vs female, or across ethnicities) is not always straightforward. For instance, the area under the receiver operating characteristic curve (AUC, also known as c-statistic) may vary across subgroups, with a model having an overall AUC of 0.85 but subgroup AUCs of 0.75 and 0.80 in male and female subgroups, respectively.

Such patterns reflect a key mathematical property of the metric itself: \textit{collapsibility}. A metric is collapsible over a variable $g$ (e.g., sex) if its marginal value equals the weighted average of the subgroup specific values, with weights determined by the marginal distribution of $g$ (e.g., male, female). When collapsibility holds, this aggregation artefact cannot occur. The phenomenon is closely related to, though not identical to Simpson’s paradox \cite{blyth1972simpson,hernan2011simpson}, whereby associations observed within subgroups disappear or reverse when data is aggregated \cite{simpson1951interpretation}. In extreme cases, marginal and conditional association metrics may have opposite directions, a pattern also referred to as the reversal paradox \cite{messick1981reversal}, aggregation bias, or the amalgamation paradox \cite{good1987amalgamation}.

In the context of performance metrics, collapsibility means that the overall value can be expressed as an appropriate weighted average of the subgroup specific values. If a metric is collapsible, the overall value $M_{\text{overall}}$ is therefore directly linked to the subgroup values $M_g$. By contrast, when a metric is non-collapsible, this relationship breaks down, making it more difficult to interpret subgroup and overall performance jointly, and potentially distorting  fairness evaluations.

This motivates a broader examination of collapsibility in the context of fairness, extending beyond the AUC example provided above. Most fairness metrics are parity based, defined as differences in subgroup specific values of model performance \cite{matos2025critical}. Applying such metrics without understanding and accounting for collapsibility can lead to misleading conclusions about subgroup and overall performance. A deeper understanding of collapsibility is therefore essential for fairness evaluation. 

The aim of this paper is to examine the collapsibility properties of 15 commonly reported performance metrics for predictive artificial intelligence (AI) models with binary outcomes, with a particular focus on the AUC.

\section{Collapsibility of Performance Metrics}
\label{sec2}

\subsection{Definition}
Let $(Y,g)$ denote samples with a binary outcome $Y\in\{0,1\}$ and a sensitive attribute $g$ (e.g., $g\in\{A,B\}$). Let $\hat p$ denote the model estimated probabilities for each sample. A performance metric $M$ is \emph{collapsible} \cite{huitfeldt2019collapsibility} over an attribute $g$ if 

\begin{align}
    M_{overall}
    &= \frac{
      \sum_{g} \left\{ w_g \times M(\hat p, Y, g) \right\}
    }{
      \sum_{g} w_g
    },
    \label{eq_collap1}
\end{align}

where $w_g$ denote the weights used to recover the overall performance value (i.e., metric specific weights). These weights are determined by the specific metric under consideration and by the way it decomposes across subgroups; for example, they may correspond to subgroup proportions or subgroup specific outcome prevalences.

A metric is collapsible over a variable $g$ if it can be written as a linear combination of its subgroup specific components. By definition, such a metric satisfies the following condition:

\begin{equation}
  \min_{g} M_g \;\;\leq\;\; M_{\text{overall}} \;\;\leq\;\; \max_{g} M_g, \label{eq:convex_bounds}
\end{equation}

meaning that overall metric values are bounded by subgroup specific values. 

If no metric specific weights exist, consistent with the decomposition of $M$, for which equation~\ref{eq_collap1} holds, then $M$ is non-collapsible with respect to $g$. In such cases, a Simpson's paradox type aggregation artefact may occur, although this is neither necessary nor sufficient for non-collapsibility. When this artefact occurs, condition~\ref{eq:convex_bounds} is violated.

Because even collapsible metrics, such as the arithmetic mean, can exhibit Simpson's paradox when subgroup comparisons are stratified by a confounding variable \cite{simpson1951interpretation}, we refer to the aforementioned phenomenon here as an ``aggregation artefact''. The term refers to situations in which the overall value $M_{\text{overall}}$ lies outside the range spanned by the subgroup values $M_g$, such that the relationship observed within subgroups is reversed or distorted after aggregation.

For non-collapsible metrics, the discrepancy between $M_{\text{overall}}$ and the weighted combination of subgroup values is captured by Jensen's gap. If no metric specific weights make the overall value equal to the weighted average of the subgroup values, then the aggregation relationship is non-linear, and may be convex or concave. For $g\in\{A,B\}$, this discrepancy can be written as

\begin{align}
    J_{\text{gap}} = w_A M_A + (1-w_A) M_B - M_{\text{overall}},
    \label{jgap}
\end{align}

where $w_A \in [0,1]$ is the weight that would recover $M_{\text{overall}}$ from $M_A$ and $M_B$, and $J_{\text{gap}} \in \mathbb{R}$.

\begin{figure}[ht]
  \centering
  \begin{tikzpicture}
    \begin{groupplot}[
      group style={group size=2 by 1, horizontal sep=1.5cm},
      width=6cm,
      height=5cm,
      xlabel={\(w_B\)},
      ylabel={$M$},
      xmin={-3/14}, xmax={17/14},
      ymin=0, ymax=1,
      domain=0:1,
      samples=200,
      variable=\pB,
      grid=both,
      xtick={0,0.5,1},
      ytick=\empty,
      clip=false,
    ]

      \nextgroupplot[title={\textit{Collapsible metric}}]
        \addplot[thick, black] {0.3 + 0.4*\pB};
        \addplot[mark=*] coordinates {(0,0.3)} node[below] {\(M_A\)};
        \draw[dashed, gray] (axis cs:{-3/14},0.3) -- (axis cs:{17/14},0.3);
        \addplot[mark=*] coordinates {(1,0.7)} node[above] {\(M_B\)};
        \draw[dashed, gray] (axis cs:{-3/14},0.7) -- (axis cs:{17/14},0.7);
        \addplot[mark=*, mark options={fill=blue}] 
          coordinates {(0.5,0.5)} node[below, blue] {\(M_{\text{overall}}\)};

      \nextgroupplot[title={\textit{Non-collapsible metric}}]
        \addplot[thick, black] {0.3 + 0.4*\pB - \pB*(1-\pB)};
        \addplot[thick, gray] {0.3 + 0.4*\pB + \pB*(1-\pB)};
        \addplot[thick, dashed, black] {0.3 + 0.4*\pB};

        \addplot[mark=*] coordinates {(0,0.3)} node[below] {\(M_A\)};
        \draw[dashed, gray] (axis cs:{-3/14},0.3) -- (axis cs:{17/14},0.3);
        \addplot[mark=*] coordinates {(1,0.7)} node[above] {\(M_B\)};
        \draw[dashed, gray] (axis cs:{-3/14},0.7) -- (axis cs:{17/14},0.7);

        \pgfmathsetmacro{\Mstar}{0.3 + 0.4*0.5}      
        \pgfmathsetmacro{\Moverall}{0.3 + 0.4*0.5 - 0.5*(1-0.5)} 

        \addplot[mark=*, mark options={fill=gray}] 
          coordinates {(0.5,\Mstar)} node[above, gray] {\(M^*\)};
        \addplot[mark=*, mark options={fill=blue}] 
          coordinates {(0.5,\Moverall)} node[below, blue] {\(M_{\text{overall}}\)};

        \draw[<->, red, thick]
          (axis cs:0.5,\Mstar) -- (axis cs:0.5,\Moverall)
          node[midway, left, font=\small, red] {\(J_{gap}\)};

    \end{groupplot}
  \end{tikzpicture}
  \caption{Collapsible vs non‐collapsible metrics as functions of a weight \(w_g\). 
  \textit{Left:} collapsible metric, \(M_{overall}\) can be recovered from linear combination of group‐specific values \(M_A\) and \(M_B\). \textit{Right:} non‐collapsible metric, curved black line (\(M_{\text{overall}}\)) deviates from the linear interpolation (\(M^*\)). This could be concave (black) or convex (grey). The vertical red arrow denotes Jensen’s gap for the concave case.}
  \label{fig:collapsible_noncollapsible}
\end{figure}
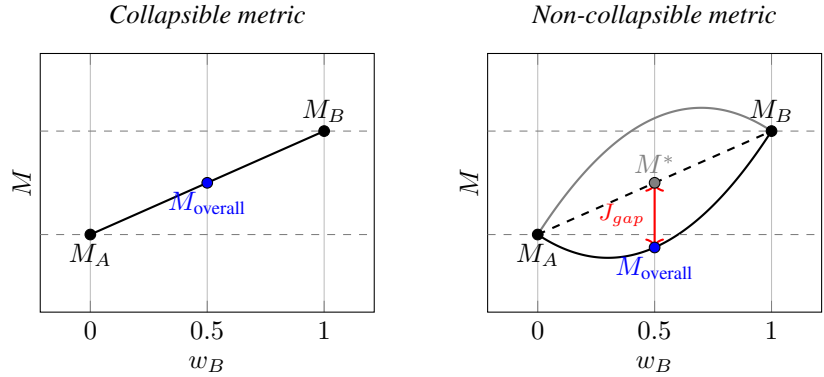

Figure~\ref{fig:collapsible_noncollapsible} illustrates this contrast, with $J_{\text{gap}} = 0$ for collapsible metrics (left), and $J_{\text{gap}} \neq 0$ for non-collapsible metrics (right).

Whilst a metric can be shown to be collapsible if Equation~\ref{eq_collap1} holds, non-collapsibility can be established by counterexample: if one can find a case in which Equation~\ref{eq:convex_bounds} is violated, then the metric is not collapsible.

\newpage
\label{sec6}

\subsection{Collapsible and non-collapsible metrics}
We examined the collapsibility of performance metrics that are considered ``recommended'' \cite{Van-Calster2024-jw,matos2025critical}, together with metrics that are widely used but not necessarily fit for purpose, such as the F1-score. These metrics are summarised in Table~\ref{tab:perf_measures}, and Table~\ref{tab:perf_measures_collapsibility} summarises their collapsibility properties, with full proofs provided in the Appendix.

Collapsibility varies across metrics depending on how their weights are computed, a property that is critical to preventing aggregation artifacts. We distinguish between \textbf{strict collapsibility}, where weights depend exclusively on subgroup properties, such as the number of events per subgroup, and \textbf{general collapsibility}, where the marginal value equals the weighted average based on any identifiable weight, such as the number of true positives according to the model.

Metrics satisfying strict collapsibility include logloss, the Brier score, accuracy, recall, specificity, and net benefit. These aggregate as linear weighted averages based on subgroup outcome prevalence ($\pi_g$) or subgroup proportion ($p^g$). 

Metrics satisfying general collapsibility include the O:E ratio, F1-score, PPV, and NPV, which remain collapsible when weighted by the model-dependent terms (e.g., true positives). 

The AUC, calibration intercept, calibration slope, expected calibration error (ECE), and Nagelkerke’s $R^2$ are non-collapsible; their overall value cannot be expressed as a weighted average of the subgroup values.

\renewcommand{\arraystretch}{1}  
\begin{table}[htbp]
\centering
\caption{Performance metrics for evaluating prediction models with binary outcomes \cite{Van-Calster2024-jw}.}
\label{tab:perf_measures}
\small
\begin{tabular}{>{\raggedright\arraybackslash}p{0.25\linewidth} >{\raggedright\arraybackslash}p{0.65\linewidth}}
\toprule
\textbf{Measure} & \textbf{Description} \\
\midrule
\multicolumn{2}{l}{\textbf{Discrimination}} \\
AUC / c-statistic & Probability that an individual with the event has a higher estimated probability than an individual without the event. \\
\midrule

\multicolumn{2}{l}{\textbf{Calibration}} \\
O:E ratio & Ratio of the number of individuals with an event to the expected number of individuals with an event according to the model. \\
Calibration intercept & Indicates whether probabilities are on average underestimated (intercept > 0), overestimated (<0), or correct (0). \\
Calibration slope & Indicates whether probabilities are too extreme (slope > 1), too modest (<1), or perfect (1). \\
Expected Calibration Error (ECE) & Weighted mean absolute difference between average probability and observed proportion from risk groups. \\
\midrule
\multicolumn{2}{l}{\textbf{Overall Performance}} \\
Logloss / Cross-entropy & Negative loglikelihood. \\
Brier score & Average squared difference between outcome labels and estimated probabilities. \\
Nagelkerke R-squared & Cox-Snell R-squared scaled (loglikelihood-based R-squared) to the value for a perfect model. \\
\midrule
\multicolumn{2}{l}{\textbf{Classification (Summary Measures)}} \\
Classification accuracy & Proportion of correctly classified individuals. \\
F1-score & Harmonic mean of precision and sensitivity. \\
\midrule
\multicolumn{2}{l}{\textbf{Classification (Partial Measures)}} \\
Sensitivity / Recall / TPR & Proportion of individuals with an event that are correctly classified. \\
Specificity / TNR & Proportion of individuals without an event that are correctly classified. \\
PPV / Precision & Proportion of high-risk individuals that have an event. \\
NPV & Proportion of low-risk individuals that do not have an event. \\
\midrule
\multicolumn{2}{l}{\textbf{Clinical Utility}} \\
Net Benefit & Net proportion of true positives, equivalent to sensitivity in the absence of false positives. \\
\bottomrule
\end{tabular}
\end{table}

\newpage
\renewcommand{\arraystretch}{1}  
\begin{table}[H]
\centering
\caption{Collapsibility properties of common performance measures.}
\label{tab:perf_measures_collapsibility}
\small
\begin{tabular}{>{\raggedright\arraybackslash}p{0.12\linewidth} >
{\raggedright\arraybackslash}p{0.19\linewidth}>
{\raggedright\arraybackslash}p{0.33\linewidth}>
{\raggedright\arraybackslash}p{0.24\linewidth}}
\toprule
\textbf{Measure} & \textbf{Collapsibility} & \textbf{Intuition} & \textbf{Weights, $\boldsymbol{w_g}$} \\
\midrule
\multicolumn{4}{l}{\textbf{Discrimination}} \\
AUC & \textbf{Non-collapsible} & Depends on both within- and cross-group pairings, which cannot be expressed as a convex combination of subgroup AUCs. & -- \\
\midrule
\multicolumn{4}{l}{\textbf{Calibration}} \\
O:E ratio & \textbf{Collapsible} (model-dependent) & Ratio of sums; expected counts decompose linearly. & $E_g / \sum_h E_h$ \\
Cal. Intercept & \textbf{Non-collapsible} & Regression coefficient, estimated globally, not an average of subgroup intercepts. & -- \\
Cal. Slope & \textbf{Non-collapsible} & Regression coefficient, estimated globally, not an average of subgroup slopes. & -- \\
ECE & \textbf{Non-collapsible} & Absolute deviations inside bins prevent linear aggregation across groups. & -- \\
\midrule
\multicolumn{4}{l}{\textbf{Overall Performance}} \\
Logloss& \textbf{Collapsible} & Log-likelihood contributions add over individuals. & $p^g$ \\
Brier score & \textbf{Collapsible} & Squared errors add over individuals. & $p^g$ \\
Nagelkerke $R^2$ & \textbf{Non-collapsible} & Nonlinear rescaling of likelihood ratio; not decomposable into group averages. & -- \\
\midrule
\multicolumn{4}{l}{\textbf{Classification (Summary Measures)}} \\
Accuracy & \textbf{Collapsible} & Correct classifications add linearly across individuals. & $p^g$ \\
F1-score & \textbf{Collapsible} (model-dependent) & Ratio of sums with group-specific denominators; can be written as weighted sum of subgroup F1. & $\dfrac{2TP_g+FP_g+FN_g}{\sum_h (2TP_h+FP_h+FN_h)}$ \\
\midrule
\multicolumn{4}{l}{\textbf{Classification (Partial Measures)}} \\
TPR & \textbf{Collapsible} & Denominator is number of positives, which decomposes via subgroup prevalences. & $\dfrac{p^g \pi_g}{\pi}$ \\
TNR & \textbf{Collapsible} & Denominator is number of negatives, which decomposes via subgroup negative prevalences. & $\dfrac{p^g (1-\pi_g)}{1-\pi}$ \\
PPV & \textbf{Collapsible} (model-dependent) & Denominator is number of predicted positives, which decomposes linearly. & $\dfrac{p^g \hat\pi_g}{\hat\pi}$ \\
NPV & \textbf{Collapsible} (model-dependent) & Denominator is number of predicted negatives, which decomposes linearly. & $\dfrac{p^g (1-\hat\pi_g)}{1-\hat\pi}$ \\
\midrule
\multicolumn{4}{l}{\textbf{Clinical Utility}} \\
Net Benefit & \textbf{Collapsible} & Defined directly in terms of collapsible counts (TP, FP). & $p^g$ \\
\bottomrule
\end{tabular}

\vspace{0.5em}
\footnotesize
$p^g = N_g/N$ is the subgroup proportion; 
$\pi_g = N^{+}_g/N_g$ is the subgroup specific prevalence of the outcome, with $\pi = N^{+}/N$ overall; 
$\hat\pi_g = \hat N^{+}_g/N_g$ is the subgroup specific predicted positive rate, with $\hat\pi = \hat N^{+}/N_{overall}$; 
$E_g = \sum_{i\in g}\hat p_i$ is the expected number of events in subgroup $g$; 
$TP_g, FP_g, FN_g$ are subgroup specific confusion matrix counts.
\end{table}

\section{AUC mixture decomposition and non-collapsibility}
\label{sec_auc}

The AUC is a commonly reported prediction model performance metric \cite{Van-Calster2024-jw,collins2024tripod+}, quantifying how well a model ranks individuals who experience the event above those who do not \cite{janssens2020reflection}. Because it evaluates all event and non-event pairs, it includes pairs that span different subgroups in fairness sensitive settings, a property previously examined through the notion of cross-AUC (xAUC) \cite{Kallus2019-yl}. The $\mathrm{xAUC_{i,j}}$ represents the probability that an individual with the event from subgroup $i$ is ranked higher than an individual without the event from subgroup $j$ \cite{Kallus2019-yl}. Both within-group and cross-group AUC are expected to influence the overall AUC. Given the central role of AUC in evaluating discrimination, we derive its decomposition under mixtures of subpopulations and formally prove that it is non-collapsible.

\subsection{Problem setting and notation}

Consider a dataset of i.i.d.\ samples $(X, G, Y)$, where $X$ denotes the predictor vector, $G$ is the subgroup indicator taking one of $K$ possible values in $\{1,\dots,K\}$ (sensitive attribute), and $Y \in \{0,1\}$ is the binary outcome. Let
\[
  p_g = P(G = g), \qquad
  p_y^g = P(G = g,\, Y = y),
\]
and define the subgroup outcome prevalence
\[
  \pi_g = P(Y = 1 \mid G = g) = \frac{p_1^g}{p_g}.
\]
By the law of total probability,
\[
  p_y = \sum_{g=1}^{K} p_y^g,
  \qquad
  p_g = p_0^g + p_1^g,
  \qquad
  p_1^g = p_g\,\pi_g,
  \quad
  p_0^g = p_g\,(1-\pi_g).
\]

A prediction model
\[
  h : \mathcal{X} \to [0,1],
  \qquad
  \hat p = h(X)
\]
produces estimated probabilities $\hat p$.

\begin{figure}[ht]
  \centering
  \begin{tikzpicture}
    \begin{groupplot}[
      group style={
        group size=2 by 1,
        horizontal sep=1.5cm,
      },
      width=6cm,
      height=5cm,
      xlabel={\(\hat p\)},
      ylabel={Density},
      xmin=0, xmax=1,
      ymin=0, ymax=6,
      domain=0:1,
      samples=200,
      variable=\phat,
      grid=both,
      ytick=\empty,
    ]
      \nextgroupplot[title={\(g=A\)}]
        \addplot[name path=f0A, thick] {40 * pow(\phat,1) * pow(1-\phat,4)};
        \addplot[name path=f1A, thick, densely dashed] {40 * pow(\phat,4) * pow(1-\phat,1)};
        \addplot[black, thin] coordinates {(0.4,0) (0.4,6)};
        \node at (axis cs:0.38,5.5)[anchor=east] {\(\tau\)};
        \addplot[pattern=north east lines, pattern color=red, draw=none, domain=0.4:1] {40 * pow(\phat,1) * pow(1-\phat,4)} \closedcycle;
        \addplot[pattern=north west lines, pattern color=black, draw=none, domain=0.4:1] {40 * pow(\phat,4) * pow(1-\phat,1)} \closedcycle;
        \node[red, font=\small, align=center] at (axis cs:0.28,1.5) {$FPR_A$};
        \node[black, font=\small, align=center] at (axis cs:0.55,3) {$TPR_A$};

      \nextgroupplot[title={\(g=B\)}]
        \addplot[name path=f0B, thick] {40 * pow(\phat,2) * pow(1-\phat,3)};
        \addplot[name path=f1B, thick, densely dashed] {40 * pow(\phat,3) * pow(1-\phat,2)};
        \addplot[black, thin] coordinates {(0.4,0) (0.4,6)};
        \node at (axis cs:0.38,5.5)[anchor=east] {\(\tau\)};
        \addplot[pattern=north east lines, pattern color=red, draw=none, domain=0.4:1] {40 * pow(\phat,2) * pow(1-\phat,3)} \closedcycle;
        \addplot[pattern=north west lines, pattern color=black, draw=none, domain=0.4:1] {40 * pow(\phat,3) * pow(1-\phat,2)} \closedcycle;
        \node[red, font=\small, align=center] at (axis cs:0.28,1.8) {$FPR_B$};
        \node[black, font=\small, align=center] at (axis cs:0.55,2.3) {$TPR_B$};

    \end{groupplot}
  \end{tikzpicture}
  \caption{Conditional estimated probability densities and shaded error‐rates for 2 groups. For both Group \(A\) and for Group \(B\), the solid line denotes the distribution \(f_0=F_0'\) (individuals without the event) and the dashed line, the distribution \(f_1=F_1'\) (individuals with the event). At a common threshold \(\tau\) (set to 0.4 in this example), the red‐hatched area shows \(\mathrm{FPR}_g\) and the black‐hatched area shows \(\mathrm{TPR}_g\) \cite{janssens2020reflection}}.
  \label{fig:group_dists}
\end{figure}
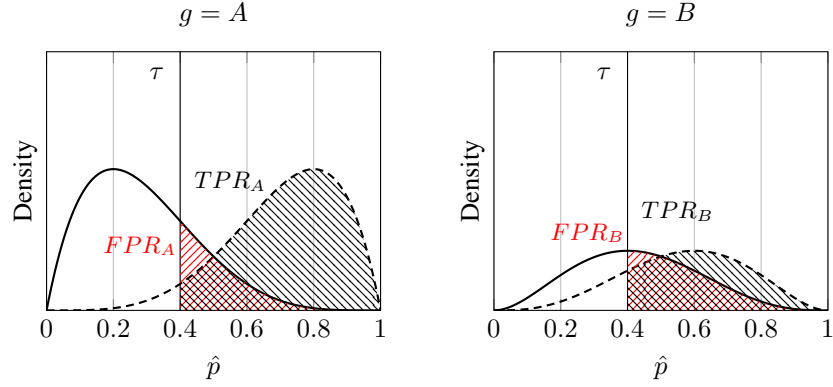

For a decision threshold $\tau \in [0,1]$, group- and class-conditional cumulative distributions can be defined as:
\[
  F_{g,y}(\tau) = P\bigl(\hat p \le \tau \mid g, Y=y\bigr),
  \quad
  g \in \{A,\dots,K\},\; y \in \{0,1\},
\]
and true positive rate (TPR) and false positive rate (FPR) functions can be defined as follows (and as illustrated in Figure \ref{fig:group_dists}):
\[
  \mathrm{TPR}_g(\tau) = 1 - F_{g,1}(\tau),
  \quad
  \mathrm{FPR}_g(\tau) = 1 - F_{g,0}(\tau),
\]

Evaluating across all subgroups, as typically done when evaluating performance at population level, yields the overall model estimated probability distributions  

\begin{align}
  F_y(t) = P(\hat p \le \tau \mid Y=y), \quad y \in \{0,1\}.
\end{align}

The AUC is then defined as \cite{gonccalves2014roc}:
\begin{align}
  \mathrm{AUC}
  = \int (1 - F_1(\tau))\,\mathrm{d}F_0(\tau).
\end{align}

\subsection{Mixture decomposition of overall AUC and non-collapsibility}

The overall AUC can also be expressed as the probability that the prediction model assigns a higher estimated probability to a randomly selected individual with the event than to a randomly selected individual without the event \cite{Faraggi2002-hm}:

\begin{equation}
  \mathrm{AUC}_{\mathrm{overall}}
  = P(\hat p_1 > \hat p_0),
\end{equation}

where $\hat p_1$ and $\hat p_0$ denote the estimated probabilities for independent draws from the event and non-event individuals, respectively.

Conditioning on subgroup membership, let $\hat p_1^i$ denote the estimated probability for an individual with the event from subgroup $i$, and let $\hat p_0^j$ denote the estimated probability for an individual without the event from subgroup $j$, where $i,j \in \{1,\dots,K\}$. Assuming independence between sampled pairs, we obtain

\begin{align}
  \mathrm{AUC}_{\mathrm{overall}}
  &= \sum_{i,j=1}^{K}
     P(G=i \mid Y=1)\,P(G=j \mid Y=0)\,
     P(\hat p_1^i > \hat p_0^j) \nonumber \\[1mm]
  &= \sum_{i,j=1}^{K}
     \frac{p_1^i}{p_1}\,\frac{p_0^j}{p_0}\,
     \mathrm{AUC}_{i,j},
     \label{eq:auc_decomposition_general}
\end{align}

This decomposition can be represented as
\begin{align}
  \mathrm{AUC}_{\mathrm{overall}}
  =
  \begin{pmatrix}
    \tfrac{p_1^{g_1}}{p_1} & \cdots & \tfrac{p_1^{g_K}}{p_1}
  \end{pmatrix}
  \mathbf{M_{\mathrm{AUC}}}
  \begin{pmatrix}
    \tfrac{p_0^{g_1}}{p_0} \\[3pt]
    \vdots \\[3pt]
    \tfrac{p_0^{g_K}}{p_0}
  \end{pmatrix},
\end{align}
where $\mathbf{M_{\mathrm{AUC}}}$ is a $K \times K$ matrix of within- and cross-group AUCs with entries $\mathrm{AUC}_{i,j} = P(\hat p_1^i > \hat p_0^j)$. The diagonal entries correspond to within-group AUCs, while the off-diagonal terms capture cross-group rankings.

\medskip

For two groups $A$ and $B$, the decomposition reduces to
\begin{align}
\boxed{
\mathrm{AUC}_{\mathrm{overall}}
=
\frac{p_1^A p_0^A}{p_1 p_0}\,\mathrm{AUC}_{A}
+\frac{p_1^A p_0^B}{p_1 p_0}\,\mathrm{xAUC}_{A,B}
+\frac{p_1^B p_0^A}{p_1 p_0}\,\mathrm{xAUC}_{B,A}
+\frac{p_1^B p_0^B}{p_1 p_0}\,\mathrm{AUC}_{B}
}
\label{mixture_2groups}
\end{align}

with

\begin{align}
\mathbf{M_{AUC}} =
\begin{pmatrix}
  \mathrm{AUC}_A      & \mathrm{xAUC}_{A,B} \\[6pt]
  \mathrm{xAUC}_{B,A} & \mathrm{AUC}_B
\end{pmatrix}.
\label{auc_matrix}
\end{align}

\medskip
In the special case where $\pi_g=\pi$, i.e., each subgroup has the same outcome prevalence, the decomposition in \eqref{eq:auc_decomposition_general} simplifies further. Substituting $p_1^i=p^i\pi$, $p_0^i=p^i(1-\pi)$, $p_1=\pi$, and $p_0=1-\pi$, we obtain
\(
  \frac{p_1^i\,p_0^j}{p_1\,p_0}
  = p^i\,p^j,
\)
which yields
\begin{align}
  \mathrm{AUC}_{\mathrm{overall}}
  = \sum_{i,j\in\{A,B\}} p^i\,p^j\,\mathrm{AUC}_{i,j}
  \label{auc_collap_same_prev}
\end{align}
and, noting that $p^B = 1 - p^A$,
\begin{align}
  \mathrm{AUC}_{\mathrm{overall}}
  = {(p^A)}^2\,\mathrm{AUC}_A
  + {(1-p^A)}^2\,\mathrm{AUC}_B
  + p^A(1-p^A)\bigl(\mathrm{xAUC}_{A,B} + \mathrm{xAUC}_{B,A}\bigr).
  \label{same_prevalences}
\end{align}

Thus, when subgroups have equal outcome prevalence, the overall AUC reduces to an average of within-group and cross-group AUC terms weighed by the subgroup proportions $p^A$.

This mixture decomposition highlights that the overall AUC depends not only on within-group AUCs but also on cross-group terms (xAUC, of equation \ref{mixture_2groups}), subgroup proportions (\(p_g\)), and subgroup specific outcome prevalences (\(\pi_g\)) \cite{sudjianto2025decomposing}. Consequently, the overall AUC may lie outside the convex bounds of the subgroup AUCs, violating the collapsibility condition \ref{eq:convex_bounds}. This property has direct implications for fairness evaluation, as discussed in Section \ref{sec_implications}.

\subsection{AUC collapsibility special cases}
\label{collapsibility_special_cases}
Despite non-collapsibility being observed in general for the AUC, there are specific scenarios when collapsibility may occur. A collapsible form of the AUC would depend solely on within-group AUC terms. Transitioning from equation~\ref{eq:auc_decomposition_general} to equation~\ref{eq_collap1} requires an appropriate combination of subgroup AUC and cross-group xAUC terms, such as the ones that we suggest and explore below.

\medskip\noindent
{\bfseries Scenario (i): \boldmath $\mathrm{xAUC}_{i,j} = \mathrm{AUC}_i$.}

Let the cross-group AUC equal the within-group AUC for each subgroup $g$:
\[
  \mathrm{xAUC}_{A,B} = \mathrm{AUC}_A,
  \quad
  \mathrm{xAUC}_{B,A} = \mathrm{AUC}_B.
\]

Applying equation~\ref{mixture_2groups} for two groups $A$ and $B$, without loss of generality, it can be shown that:

\begin{align}
  \mathrm{AUC}_{\mathrm{overall}}
  &= \frac{p_1^A p_0^A}{p_1 p_0}\,\mathrm{AUC}_A
   + \frac{p_1^A p_0^B}{p_1 p_0}\,\mathrm{xAUC}_{A,B}
   + \frac{p_1^B p_0^A}{p_1 p_0}\,\mathrm{xAUC}_{B,A}
   + \frac{p_1^B p_0^B}{p_1 p_0}\,\mathrm{AUC}_B\notag\\[4pt]
  &= \frac{p_1^A (p_0^A + p_0^B)}{p_1 p_0}\,\mathrm{AUC}_A
   + \frac{p_1^B (p_0^A + p_0^B)}{p_1 p_0}\,\mathrm{AUC}_B \notag\\[4pt]
  &= \sum_{g} \frac{\pi_g}{\pi}\,p^g\,\mathrm{AUC}_g.
  \label{eq:non-collap}
\end{align}

Thus, if cross-group AUC values equal the within-group AUC values, i.e., individuals without the event are identically distributed across subgroups, then $\mathrm{AUC}_{\mathrm{overall}}$ can be written as a linear combination, achieved with \(w_g=\frac{\pi_g}{\pi}\,p^g\) (\ref{eq_collap1}). 


\medskip\noindent
{\bfseries Scenario (ii): \boldmath $\mathrm{xAUC}_{i,j} = \mathrm{AUC}_i$ and $\pi_g=\pi$.}

In addition to the cross-group AUCs equalling the within-group AUCs, suppose that the subgroup outcome prevalences also match (i.e., \(\pi_g = \pi\)). Then, $\mathrm{AUC}_{\mathrm{overall}}$ can be expressed as:

\begin{align}
  \mathrm{AUC}_{\mathrm{overall}}
  &= \sum_{g} \,p^g\,\mathrm{AUC}_g,
  \label{eq:colla_same_prev}
\end{align}

These special cases are presented here as theoretical exceptions, and in practice the AUC should generally be treated as non-collapsible in the population of interest. In Section \ref{sec_implications}, we explore scenarios in which $\mathrm{xAUC}_{i,j} \neq \mathrm{AUC}_i$, and examine how these may lead to counter-intuitive fairness evaluations. This raises deeper questions about the AUC itself, particularly its sensitivity to case-mix differences and its relationship to performance metrics beyond discrimination, such as calibration.

\subsection{AUC estimation under normal assumptions}
To support the forthcoming examples, we derive the closed-form expression for the AUC when the overall distribution arises from a mixture of subpopulations (equation \ref{eq:auc_decomposition_general}), under normality assumptions.

There are multiple ways of estimating the AUC, including the Mann--Whitney statistic, kernel smoothing, normal assumptions, and empirical transformations to normality \cite{Faraggi2002-hm,zhou2009statistical}. Here, for illustrative purposes, we derive a closed-form expression for the overall AUC as defined by equation~\ref{eq:auc_decomposition_general}, under a truncated normal approximation. Without loss of generality, suppose that within each group \(g\in\{A,B\}\), the estimated probabilities for those with and without the event are approximately truncated normal on \([0,1]\):
\begin{align}
  \hat p_0^A &\sim \mathcal{TN}_{[0,1]}(\mu_{0}^A,\;(\sigma_{0}^A)^2),\notag
  \qquad
  \hat p_1^A \sim \mathcal{TN}_{[0,1]}(\mu_{1}^A,\;(\sigma_{1}^A)^2),\notag
  \\
  \hat p_0^B &\sim \mathcal{TN}_{[0,1]}(\mu_{0}^B,\;(\sigma_{0}^B)^2),\notag
  \qquad
  \hat p_1^B \sim \mathcal{TN}_{[0,1]}(\mu_{1}^B,\;(\sigma_{1}^B)^2).\notag
\end{align}

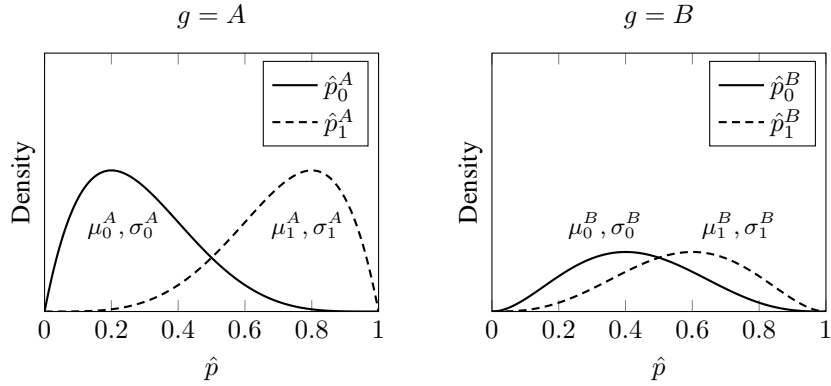
\begin{figure}[ht]
  \centering
  \begin{tikzpicture}
    \begin{groupplot}[
      group style={
        group size=2 by 1,
        horizontal sep=1.5cm,
      },
      width=6cm,
      height=5cm,
      xlabel={\(\hat p\)},
      ylabel={Density},
      xmin=0, xmax=1,
      ymin=0, ymax=6,
      domain=0:1,
      samples=200,
      variable=\phat,
      legend pos=north east,
      legend cell align=left,
      ytick=\empty,
    ]
      \nextgroupplot[title={\(g=A\)}]
        \addplot[thick] {40 * pow(\phat,1) * pow(1-\phat,4)};
        \addlegendentry{\(\hat{p}_0^A\)}
        \addplot[thick,densely dashed] {40 * pow(\phat,4) * pow(1-\phat,1)};
        \addlegendentry{\(\hat{p}_1^A\)}

        \node[gray, font=\small, anchor=south] at (axis cs:0.2857,6) {\(\mu_{0}^A\)};
        \node[gray, font=\small, anchor=south] at (axis cs:0.7143,6) {\(\mu_{1}^A\)};

        \node[black, font=\small, anchor=west] at (axis cs:0.1,2.0) {\(\mu_{0}^A,
        \sigma_{0}^A\)};
        \node[black, font=\small, anchor=west] at (axis cs:0.65,2.0) {\(\mu_{1}^A,
        \sigma_{1}^A\)};

      \nextgroupplot[title={\(g=B\)}]
        \addplot[thick] {40 * pow(\phat,2) * pow(1-\phat,3)};
        \addlegendentry{\(\hat{p}_0^B\)}
        \addplot[thick,densely dashed] {40 * pow(\phat,3) * pow(1-\phat,2)};
        \addlegendentry{\(\hat{p}_1^B\)}

        \node[black, font=\small, anchor=west] at (axis cs:0.20,2.0) {\(\mu_{0}^B,\sigma_{0}^B\)};
        \node[black, font=\small, anchor=west] at (axis cs:0.60,2.0) {\(\mu_{1}^B,\sigma_{1}^B\)};

    \end{groupplot}
  \end{tikzpicture}
  \caption{Group‐conditional densities for those without the event (\(p_0^g\), solid) and those with the event (\(p_1^g\), dashed). Nearby text labels indicate the corresponding mean and standard deviation \(\mu_{y}^g,\sigma_{y}^g\).}
  \label{fig:group_dists_mu_sigma}
\end{figure}

Then for any pair \((i,j)\) the component AUC admits the closed-form:
\begin{align}
  \mathrm{AUC}_{i,j}
  = P\bigl(\hat p_1^i>\hat p_0^j\bigr)
  = \Phi\!\Bigl(\frac{a_{ij}}{\sqrt{1 + b_{ij}^{2}}}\Bigr),
  \label{auc_comp_closed_form}
\end{align}
where $a$ is the separation coefficient and $b$ the symmetry coefficient,
\[
  a_{ij}
  = \frac{\mu_{1}^i - \mu_{0}^j}{\sigma_{1}^i},
  \qquad
  b_{ij}
  = \frac{\sigma_{0}^j}{\sigma_{1}^i}.
\]
\medskip
By the mixture decomposition ~\ref{eq:auc_decomposition_general}, the overall AUC becomes
\begin{align}
  \mathrm{AUC}_{\mathrm{overall}}
  = \sum_{i,j\in\{A,B\}}
      \frac{p_1^i}{p_1}\,\frac{p_0^j}{p_0}\,
      \Phi\!\Bigl(\frac{a_{ij}}{\sqrt{1 + b_{ij}^{2}}}\Bigr)
      \label{estimation_mixture}
\end{align}

Although assuming that the model’s estimated probabilities follow a normal distribution may not accurately reflect real-world scenarios, we adopt this assumption in what follows for illustration and interpretability.

\section{AUC non-collapsibility and fairness implications}
\label{sec_implications}

Non-collapsibility can be a practical concern for fairness assessments. A collapsible metric always remains within the range defined by the subgroup minimum and maximum (equation~\ref{eq:convex_bounds}); in contrast, a non-collapsible metric, such as the AUC, can breach these bounds, manifesting as an aggregation artefact (examples in Figure~\ref{fig:combined_auc}). Collapsible metrics also permit straightforward reasoning from subgroup to overall level. For the AUC in particular, the overall value entangles within-group and cross-group pairwise comparisons, which complicates interpretation and communication among statisticians, clinicians, and policymakers. This section examines specific ways in which AUC non-collapsibility can affect fairness evaluation.

\subsection{Counter-intuitive fairness scenarios}

\begin{figure}[H]
  \centering

    \begin{subfigure}[t]{0.9\textwidth}
    \centering
    \begin{tikzpicture}[>=stealth, scale=1]
      \draw[->, thick] (0,0) -- (9,0);

      \draw (5,0.08) -- (5,-0.08) node[below] {AUC$_A$ = AUC$_B$};

      \node[draw, rectangle, minimum width=1.6cm, minimum height=0.6cm] (b1) at (1,1) {AUC$_\mathrm{overall}$};
      \draw[->] (1,0.6) -- (1,0.18);
      \node[right] at (1,0.45) {?};

      \node[anchor=south west] at (0,1.6) {\textbf{(a) Subgroup parity satisfied, but overall AUC is lower:}};
    \end{tikzpicture}
  \end{subfigure}

  \vspace{0.5cm} 


  \begin{subfigure}[t]{0.9\textwidth}
    \centering
    \begin{tikzpicture}[>=stealth, scale=1]
      \draw[->, thick] (0,0) -- (9,0);

      \draw (3,0.08) -- (3,-0.08) node[below] {AUC$_B$};
      \draw (7,0.08) -- (7,-0.08) node[below] {AUC$_A$};

      \node[draw, rectangle, minimum width=1.6cm, minimum height=0.6cm] (b3) at (3,1) {AUC$_\mathrm{overall}$};
      \draw[->] (3,0.6) -- (3,0.18);
      \node[right] at (3,0.45) {?};

      \node[anchor=south west] at (0,1.6) {\textbf{(b) AUC$_B$ and AUC$_\mathrm{overall}$ match, but AUC$_A$ is higher:}};
    \end{tikzpicture}
  \end{subfigure}

  \caption{Two scenarios illustrating relationships between subgroup AUCs and the overall AUC.}
  \label{fig:combined_auc_comparison_vertical}
\end{figure}
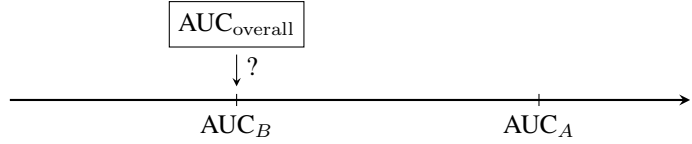

Jensen’s gap, \(J_{gap}\), arises from the non-collapsibility of the AUC and quantifies the deviation of the overall metric from the linear combination of subgroup metrics. Here we use \(J_{gap}\) to illustrate and examine two counter-intuitive, yet plausible, fairness scenarios (Figure~\ref{fig:combined_auc_comparison_vertical}).

\medskip\noindent
{\bfseries Scenario (i): \boldmath $\mathrm{AUC}_{A} = \mathrm{AUC}_B$.}

\begin{figure}[htbp]
\centering

\begin{subfigure}[t]{0.9\linewidth}
    \centering
    \includegraphics[width=\linewidth]{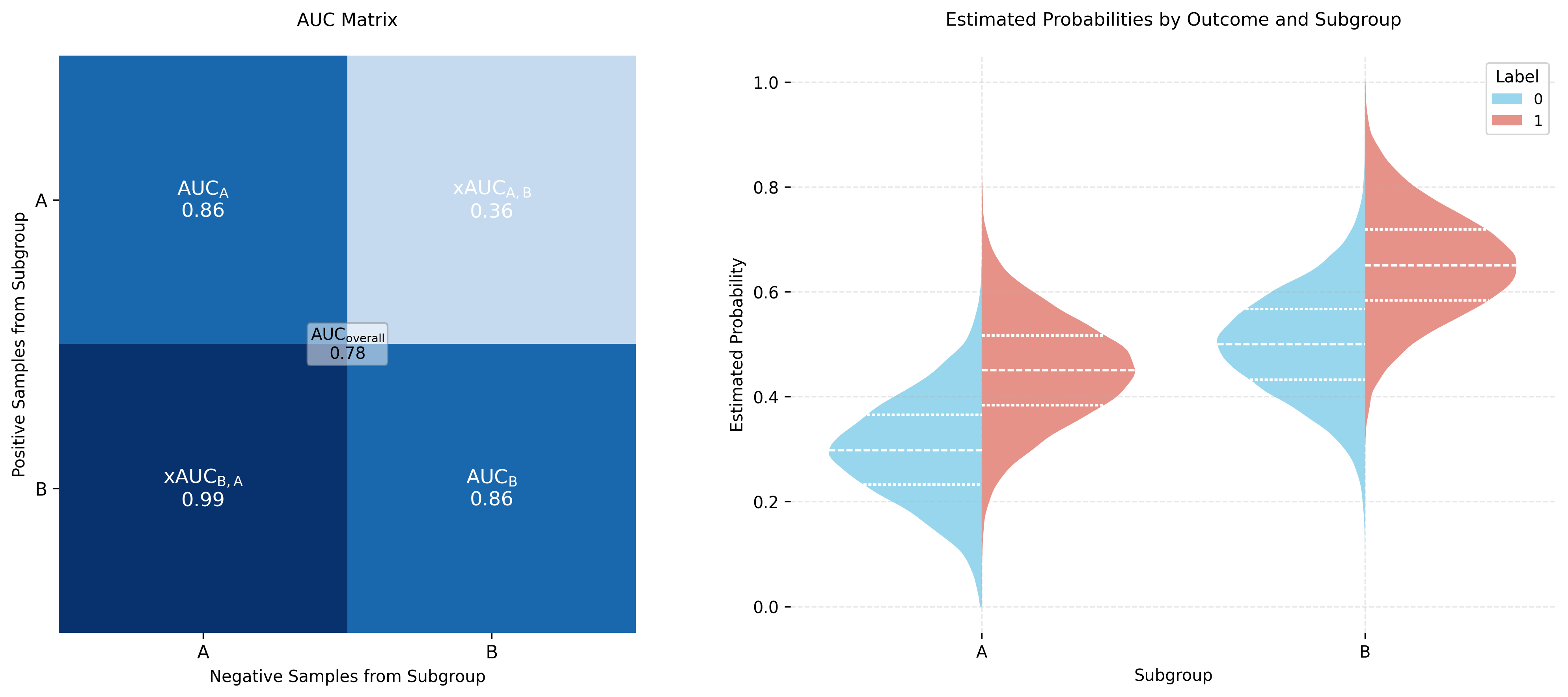}
    \caption{Simulated scenario in which subgroup parity is satisfied ($\mathrm{AUC_A} = \mathrm{AUC_B}$) but overall AUC is lower than both subgroup AUC values. Estimated probability distributions follow $\mu_0^A = 0.3$, $\mu_1^A = 0.45$, $\mu_0^B = 0.5$, and $\mu_1^B = 0.65$.}
    \label{fig:disagg_auc1}
\end{subfigure}

\vspace{1em} 

\begin{subfigure}[t]{0.9\linewidth}
    \centering
    \includegraphics[width=\linewidth]{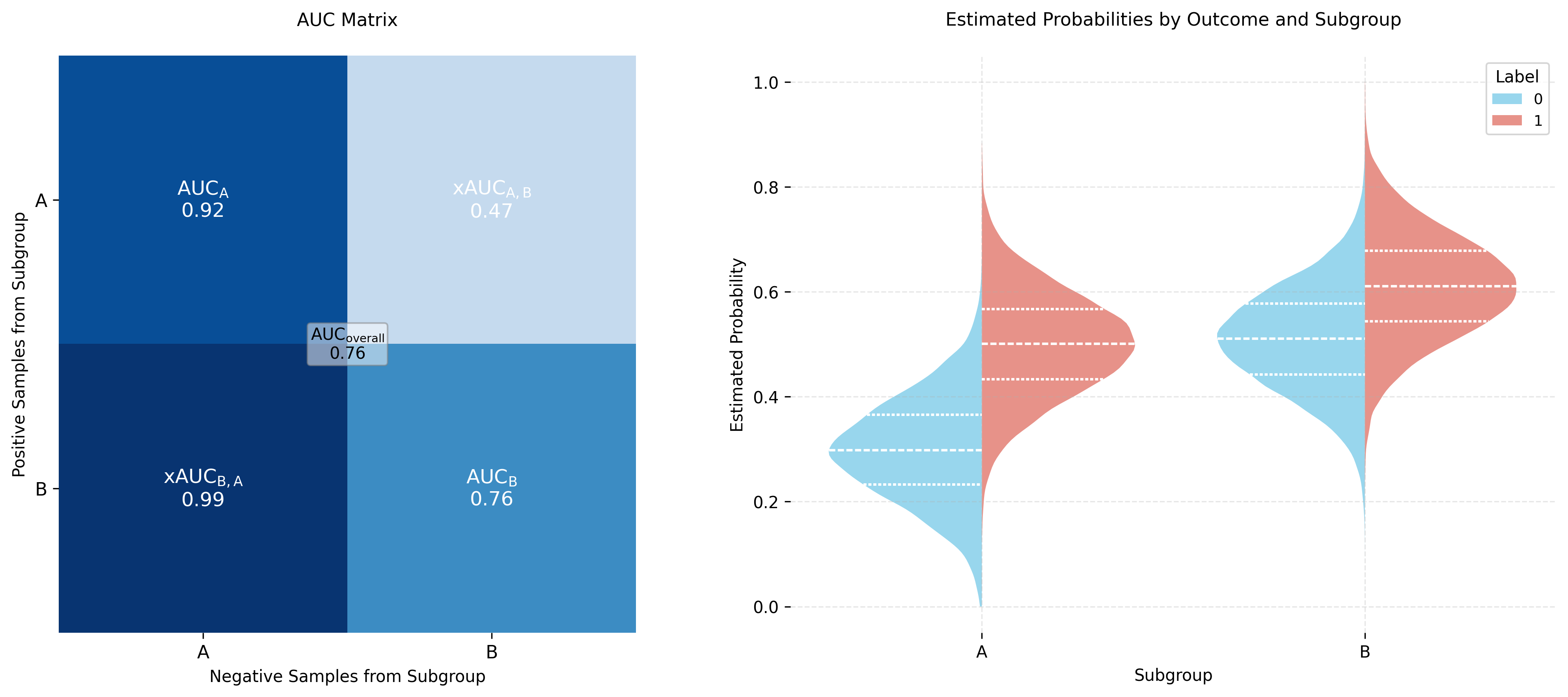}
    \caption{Simulated scenario in which $\mathrm{AUC_{B}} = \mathrm{AUC_{overall}}$, but $\mathrm{AUC_{A}}$ falls outside the range. Estimated probability distributions follow $\mu_0^A = 0.3$, $\mu_1^A = 0.5$, $\mu_0^B = 0.51$, and $\mu_1^B = 0.61$.}
    \label{fig:fig4}
\end{subfigure}

\caption{Illustrative cases comparing subgroup and overall AUC relationships. On the left panel, AUC component values are shown in the form of a matrix, $\mathrm{M_{AUC}}$, with $\mathrm{AUC_{overall}}$ for the pooled sample in the centre. On the right panel, the distributions of the estimated probability by outcome and subgroup are shown. Within each subgroup $g \in \{A,B\}$ the estimated probabilities for non-events and events are drawn from truncated normal distributions on $[0,1]$, using 100,000 samples in total. For both scenarios, $p^A = 0.3$, $\pi^A = \pi^B = \pi = 0.5$, and the estimated probabilities follow $\sigma = 0.1$ for all strata. The parameters defined here are common to both scenarios but $\mu_0^A$, $\mu_1^A$, $\mu_0^B$, $\mu_1^B$ are specific to each scenario above.}
\label{fig:combined_auc}
\end{figure}

Figure~\ref{fig:disagg_auc1} shows that subgroup parity may hold (\(AUC_A = 0.86\) and \(AUC_B = 0.86\)) while the overall AUC is substantially lower (\(AUC_{\text{overall}} = 0.78\)). This apparent paradox is explained by the cross-group AUC terms: \(\mathrm{xAUC}_{A,B}\) is disproportionately low (0.36), whereas \(\mathrm{xAUC}_{B,A}\) is very high (0.99), leading to an overall AUC that is lower than both subgroups' (similar) performance. Mathematically, when subgroup parity is satisfied, i.e.,\ $AUC_A = AUC_B$:

\begin{align}
AUC_A - AUC_{\text{overall}}
&= AUC_A - \big( w_A AUC_A + w_B AUC_B - J_{gap}\big) \notag\\ 
&=  AUC_A - \big(w_A AUC_A + (1-w_A) AUC_A - J_{gap} \big) = J_{gap}.
\label{scenario_a}
\end{align}

\medskip\noindent
{\bfseries Scenario (ii): \boldmath $\mathrm{AUC}_{overall} = \mathrm{AUC}_B$.}

Figure~\ref{fig:fig4} illustrates a setting in which one subgroup’s AUC (\(AUC_B = 0.76\)) is similar to the overall AUC (\(AUC_{\text{overall}} = 0.76\)) but the other subgroup AUC (\(AUC_A = 0.92\)) is markedly higher. Once again, the difference arises from the cross-group AUC terms, which are further weighted by subgroup proportions and subgroup outcome prevalence differences. Formally, when one subgroup metric equals the overall metric, i.e., $AUC_A = AUC_{\mathrm{overall}}$:

\begin{align}
AUC_A - AUC_B 
&= \frac{AUC_A - AUC_{\mathrm{overall}} + S}{w_B} = \frac{J_{gap}}{w_B}.
\label{scenario_b}
\end{align}

These examples illustrate that the counter-intuitive behaviours are directly tied to the non-collapsibility of the AUC, as captured by $J_{gap}$. The resulting differences are not artefacts of estimation but rather consequences of the decomposition in equation~\ref{eq:auc_decomposition_general}. Examining the distribution of the estimated probabilities generated by the model in both scenarios reveals substantial differences between groups $A$ and $B$ that account for the observed AUC values. The following section investigates these distributions, why $J_{gap}$ may be non-zero, and explores how its magnitude reflects the interaction between case-mix differences and other dimensions of model performance.

An interactive web application for exploring the proposed concepts and the mentioned scenarios is available at \url{https://collapsibility.streamlit.app/}. The corresponding Python code, used to implement the app and produce all demonstrations presented in this work, is available at \url{https://github.com/joamats/collapsibility}.

\subsection{Drivers of non-collapsible behaviour}

To understand why the AUC may fail to be collapsible, and how this leads to the counter-intuitive scenarios above, we examine how differences in the distributions of the estimated model probabilities across subgroups influence both the within-group AUC and the cross-group xAUC terms.

For illustration, suppose that there are two groups, \(A\) and \(B\), defined by a sensitive attribute \(g\), and assume normality (truncated to \([0,1]\), by definition) of the estimated probabilities with equal standard deviations across all subgroup-outcome strata \cite{austin2012interpreting}:
\[
  \sigma_{0}^{A} = \sigma_{1}^{A} = \sigma_{0}^{B} = \sigma_{1}^{B}.
\]
Under this assumption the symmetry coefficient \(b_{ij}\) appearing in equation~\ref{auc_comp_closed_form} becomes \(b_{ij}=1\). The component AUC therefore simplifies to
\begin{align}
  \mathrm{AUC}_{i,j}
  = \Phi\!\left(\frac{a_{ij}}{\sqrt{2}}\right)
  = \Phi\!\left(\frac{\mu_{1}^{i}-\mu_{0}^{j}}{\sqrt{2}\,\sigma}\right),
\end{align}
where \(\sigma\) denotes the common standard deviation.

As noted in subsection~\ref{collapsibility_special_cases}, collapsibility can be achieved via equality between certain within-group and cross-group AUC terms, such as \(\mathrm{AUC}_i = \mathrm{xAUC}_{i,j}\). Under the simplified expression above, this condition becomes
\[
  \mu_{1}^{i} - \mu_{0}^{i}
  = \mu_{1}^{i} - \mu_{0}^{j}
  \quad\Longleftrightarrow\quad
  \mu_{0}^{i} = \mu_{0}^{j}.
\]
Alternatively, if collapsibility were to arise through \(\mathrm{AUC}_j = \mathrm{xAUC}_{i,j}\), then
\[
  \mu_{1}^{i} - \mu_{0}^{j}
  = \mu_{1}^{j} - \mu_{0}^{j}
  \quad\Longleftrightarrow\quad
  \mu_{1}^{i} = \mu_{1}^{j}.
\]
Hence deviations from collapsibility can be attributable to differences in the mean estimated probabilities across subgroups, specifically to contrasts of the form \(\mu_{1}^{A}-\mu_{1}^{B}\) and \(\mu_{0}^{A}-\mu_{0}^{B}\). The quantity \(\mu_{1}^{A}-\mu_{1}^{B}\) has been referred to in the literature as ``balance for the positive class''  and \(\mu_{0}^{A}-\mu_{0}^{B}\) as
``balance for the negative class'' \cite{kleinberg2016inherent}.

This raises the question of why the means of the estimated probabilities should differ between subgroups at all. Several mechanisms may lead to such differences: two that are common in a fairness evaluation are highlighted below.

\medskip
\noindent\textbf{Driver (i): Differences in subgroup outcome prevalence under fair mean calibration.}\\
Assume the prediction model is well-calibrated within each subgroup in the sense of calibration-in-the-large, meaning the average estimated probability matches the outcome prevalence \cite{van2016calibration}. When within-group discrimination is identical, differences in subgroup outcome prevalence \(\pi_A\) and \(\pi_B\) will induce corresponding differences in the mean estimated probabilities, since a group with a higher outcome prevalence will, on average, have higher estimated probabilities. These shifts in \(\mu_{1}^{g}\) and \(\mu_{0}^{g}\) propagate into the xAUC terms, potentially creating cross-group asymmetries even when the within-group AUCs remain comparable. Under fairness evaluation involving two or more heterogeneous subpopulations, such xAUC differences may have counter-intuitive effects on the overall AUC.

\medskip
\noindent\textbf{Driver (ii): Differences in subgroup calibration under equal subgroup outcome prevalence.}\\
Now suppose the subgroups share a common outcome prevalence but differ in their mean calibration. For instance, systematic over- or underestimation in one subgroup would shift either \(\mu_{1}^{g}\) or \(\mu_{0}^{g}\) relative to the proportion of observed outcomes. Such differences again alter \(\mathrm{xAUC}_{i,j}\) even when the within-group discrimination remains similar.

These two mechanisms motivate a closer examination of how case-mix differences and miscalibration differences respectively contribute to AUC behaviour, as detailed below.

\subsubsection{The role of case-mix differences toward naturally occurring subgroup differences}

Differences in AUC across subgroups do not necessarily stem from shortcomings in the model's ability to discriminate, nor do they necessarily imply a lack of fairness.  They may instead reflect differences in the underlying case-mix \cite{vergouwe2010external}, a phenomenon broadly referred to as spectrum bias \cite{tseng2021spectrum}, whereby the performance of a prediction model varies when applied to populations with differing distributions of patient characteristics.

Here we focus on two quantities that can vary across subgroups defined by a sensitive attribute and affect the mixture decomposition of the overall AUC: subgroup proportions and subgroup specific outcome prevalences. Subgroup proportions reflect population composition, whereas differences in outcome prevalence may arise from underlying case-mix differences in baseline characteristics. Both influence the mixture decomposition of the overall AUC and thereby the magnitude of non-collapsibility effects.

Let \(p^g\) denote the proportion of individuals in subgroup \(g\), and \(\pi_g\) the outcome prevalence within that subgroup. Differences in these quantities influence the mixture decomposition of the overall AUC and thereby contribute to the magnitude of non-collapsibility effects.

\medskip
\noindent\textbf{(i) Different subgroup proportions:}\\
Subgroup proportions \(p^g\) do not affect the AUC or xAUC terms, but they do determine the relative influence of these components in the overall mixture expression (equation~\ref{estimation_mixture}). Consequently, even if a model behaves identically across subgroups, different proportions of the subgroups can impact on the overall AUC.

\medskip
\noindent\textbf{(ii) Different subgroup specific outcome prevalences:}\\
Although the AUC is invariant to outcome prevalence in a homogeneous population, differences in \(\pi_g\) across subgroups influence the mixture decomposition in two important ways. First, as discussed in Section~\ref{collapsibility_special_cases}, varying \(\pi_g\) affects the means of the estimated probability distributions -- potentially to different extents, depending on how much subgroup specific calibration-in-the-large follows such differences or not. These mean shifts alter the relative positions of the subgroup estimated probability distributions and therefore modify the xAUC terms, even when the within-group discrimination remains unchanged. Second, subgroup outcome prevalences determine the mixture weights \(\frac{p_1^g}{p_1}\) and \(\frac{p_0^g}{p_0}\), which govern the contribution of each subgroup-outcome stratum to the overall AUC. Differences in \(\pi_g\) therefore amplify or attenuate the influence of particular strata on the mixture.

\medskip

\begin{figure}[htbp]
    \centering
    \includegraphics[width=.8\linewidth]{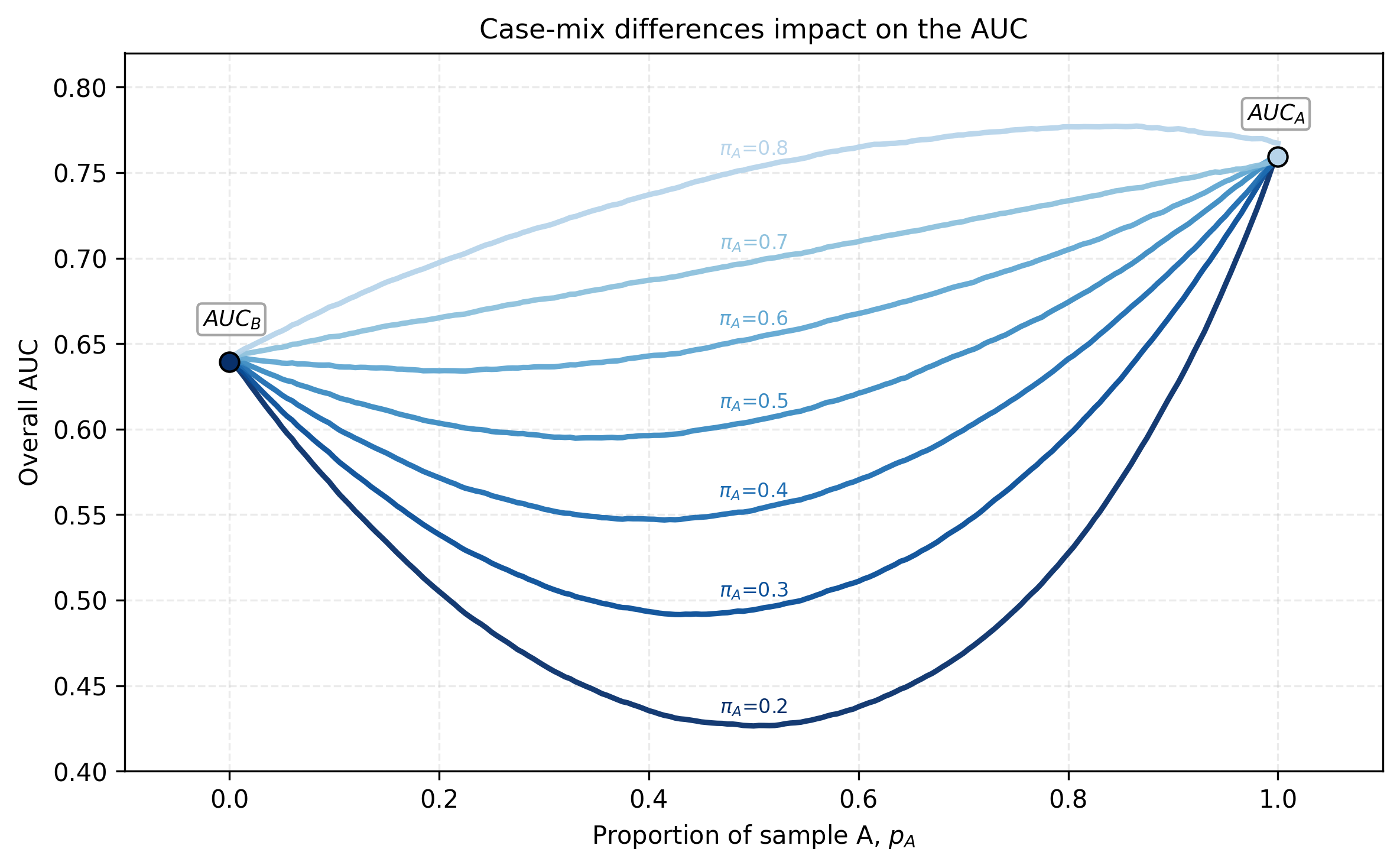}

\caption{\small Overall AUC of a single model applied to subgroups A and B as the sample proportion $p_A$ varies from 0 to 1. Curves show different prevalences in A, $\pi_A\in\{0.2,0.3,\dots,0.8\}$, with fixed $\pi_B=0.5$. There are 20{,}000 samples in total for each scenario. For each subgroup $g\in\{A,B\}$ estimated probabilities for non-events and events were drawn from truncated Normal$([0,1])$ with means $(\mu_0^A,\mu_1^A)=(0.6,0.7)$ and $(\mu_0^B,\mu_1^B)=(0.25,0.30)$, and fixed $\sigma=0.1$. Endpoints $p_A=0,1$ recover $\mathrm{AUC_B}$ and $\mathrm{AUC_A}$, respectively.}

    \label{fig:fig8}
\end{figure}

Figure~\ref{fig:fig8} illustrates these effects, showing how variations in subgroup proportions and subgroup specific prevalences jointly influence the overall AUC under a fixed underlying model. These observations highlight that case-mix differences, including variations in subgroup proportions and subgroup specific outcome prevalences, can produce non-collapsible behaviour that may inflate ``fairness gaps'' that do not necessarily translate to unfairness in the model. Rather, such differences may reflect population heterogeneity and should be interpreted with care in fairness evaluations.

This perspective aligns closely with earlier work by van Klaveren and colleagues, who proposed a concordance based metric designed to disentangle the contribution of case-mix heterogeneity from true differences in discriminative ability \cite{van2016new}. It is also consistent with propensity based standardisation approaches, which aim to disentangle differences in discrimination due to case-mix from differences due to model generalisability across samples \cite{de2023propensity}. Furthermore, Vergouwe and colleagues demonstrate that shifts in the distribution of patient characteristics can affect model performance, even when the underlying model remains unchanged \cite{vergouwe2010external}. Notably, they also document a scenario in which more severe case-mix changes induce systematic miscalibration, which naturally raises the question of how calibration and discrimination interact. We therefore now examine the role of miscalibration differences in shaping observed AUC behaviour.

\subsubsection{The role of miscalibration differences}

The AUC is a rank based metric, invariant to any strictly monotone transformation of the estimated probabilities, and is therefore generally, considered independent of calibration \cite{martens2016small}. This interpretation warrants some clarification. Pencina and colleagues have noted that miscalibration can affect the empirical AUC in practice \cite{pencina2017discrimination}. More recently, Sadatsafavi and colleagues introduced a ``model-based ROC curve'', which separates the effects of case-mix heterogeneity and miscalibration \cite{sadatsafavi2022model}. These contributions motivate a closer examination of how miscalibration interacts with case-mix to affect discrimination, particularly in fairness contexts where subgroup differences are of key interest.

In heterogeneous populations, the overall AUC reflects not only within-group discrimination but also cross-group estimated probability comparisons. Consequently, even equal within-group AUCs can yield a higher, or lower, overall AUC if groups differ in mean calibration or outcome prevalence. This conflation complicates fairness evaluations, since variations in overall AUC may arise from calibration discrepancies across subgroups, rather than genuine differences in discrimination. The form of calibration relevant here corresponds to calibration at the level of covariate patterns, whereby estimated probabilities match observed event rates for each subgroup specific covariate pattern. As discussed by Van Calster and colleagues \cite{van2016calibration}, this ``strong calibration'' is rarely achievable, yet systematic differences in estimated probabilities propagate directly into the AUC decomposition. These considerations further underscore the importance of evaluating discrimination and calibration jointly \cite{Van-Calster2024-jw}, including subgroup performance alongside overall performance.

\newpage
\subsection{Clinical illustration: EuroSCORE II}

Here, we use the European System for Cardiac Operative Risk Evaluation (EuroSCORE) II, a clinical prediction model for in-hospital mortality after cardiac surgery \cite{nashef2012euroscore}, as the setting for three hypothetical fairness evaluations stratified by sex. The numerical values are illustrative and are not derived from an empirical analysis of EuroSCORE II. Box~\ref{box:euroscore_fairness} illustrates how differences in overall, subgroup, and cross-group AUCs may arise without necessarily indicating model unfairness.

\medskip
\begin{discussion_box}[Illustrative fairness evaluation scenarios using EuroSCORE II]\label{box:euroscore_fairness}
\small

\textbf{Apparent subgroup disparity:}
Assume that EuroSCORE II has higher discrimination among male patients than among female patients ($\mathrm{AUC}_{M}=0.88$ versus $\mathrm{AUC}_{F}=0.73$), while $\mathrm{AUC}_{\mathrm{global}}=0.85$. The cross-group discrimination values are also asymmetric, with $\mathrm{xAUC}_{M,F}=0.67$ and $\mathrm{xAUC}_{F,M}=0.92$. Here, $xAUC_{M,F}=0.67$ indicates that a male patient who dies is ``ranked'' above a female patient who survives only 67\% of the time, whereas $xAUC_{F,M}=0.92$ indicates that a female patient who dies is ``ranked'' above a male patient who survives 92\% of the time. These within- and cross-group differences identify potential fairness concerns that warrant further investigation, but they do not by themselves establish that the model is unfair.

\medskip
\medskip

\textbf{Improved subgroup performance but lower overall AUC:}
Discrimination among female patients increases to $\mathrm{AUC}_F = 0.80$, while discrimination among male patients remains $\mathrm{AUC}_M = 0.88$. Nevertheless, the overall AUC decreases from 0.85 to 0.84. The corresponding cross-group values are $\mathrm{xAUC}_{M,F} = 0.44$ and $\mathrm{xAUC}_{F,M} = 0.99$. While $\mathrm{xAUC}_{F,M}=0.99$ indicates almost perfect ranking of female patients who die above male patients who survive, $\mathrm{xAUC}_{M,F}=0.44$ indicates that male patients who die are ranked above female survivors less often than expected by chance. The lower overall AUC should therefore not be interpreted as evidence that the model has become worse overall, as it reflects changes in cross-group rankings rather than within-group performance alone.

\medskip
\medskip

\textbf{Subgroup parity without improvement in overall AUC:}

Within-group discrimination is equalised ($\mathrm{AUC}_M = \mathrm{AUC}_F = 0.86$), while $\mathrm{AUC}_{\mathrm{global}}$ remains 0.84. However, the cross-group AUCs remain markedly different, with $\mathrm{xAUC}_{M,F} = 0.53$ and $\mathrm{xAUC}_{F,M} = 0.98$, despite identical subgroup AUCs. Specifically, $xAUC_{M,F}=0.53$ indicates that male patients who die are only slightly more likely than chance to be ``ranked'' above female survivors, whereas $xAUC_{F,M}=0.98$ indicates near-perfect ranking of female patients who die above male survivors. Equality of subgroup AUCs therefore does not imply equality across all dimensions of discrimination fairness.

\end{discussion_box}

\newpage
\section{Discussion}

\subsection{Key finding: Non-collapsibility matters for fairness checks}

When a metric is non-collapsible, several interpretation complications arise. The relationship between subgroup and overall performance can become obscured, as overall metrics can fail to reflect subgroup behaviour faithfully or to act as a linear combination of their constituent parts. This mismatch can create natural inconsistencies between subgroup level and population level results, potentially undermining confidence in fairness evaluations.

In some settings, non-collapsibility can also introduce spurious disparities between subgroups and overall estimates, generating differences driven by subgroup proportions or subgroup outcome prevalence, rather than by genuine model behaviour. It can also lead to counter-intuitive consequences for fairness interventions, since improvements in subgroup metrics do not necessarily translate into, and may even diminish, overall performance. Finally, because the overall AUC includes both within-group and cross-group pairwise comparisons, it may conflate discrimination with calibration differences across subgroups, further complicating interpretation in heterogeneous populations.

\subsection{Recommendations}
The findings presented in this paper suggest several recommendations for researchers and practitioners evaluating clinical prediction models, particularly in fairness contexts (Box~\ref{box:first}). First, before conducting subgroup analyses, it is important to establish whether the metrics of interest are collapsible. When a metric is non-collapsible, discrepancies between subgroup level and overall performance may arise naturally, even when the model behaves consistently across groups, due to case-mix differences. Second, metric estimates should always be interpreted in relation to subgroup proportions, subgroup outcome prevalences, and complementary performance domains, such as calibration performance when examining the AUC, and vice-versa. Third, transparent reporting of subgroup composition and outcome distributions is therefore essential for evaluating both model performance and fairness, and adherence to TRIPOD+AI remains fundamental \cite{collins2024tripod+}. Finally, sufficient sample size is essential for subgroup analyses \cite{riley2025decomposition1, riley2025decomposition2, whittle2026decomposition3}, as even collapsible metrics are subject to sampling variability and small samples can lead to chance differences between subgroup and overall performance that are difficult to interpret.

\medskip
\begin{discussion_box}[Key considerations and recommendations]\label{box:first}
\small

      Collapsibility properties of performance metrics should be considered when conducting subgroup analyses of performance of predictive AI. 

      \medskip
      \medskip

     The AUC, calibration intercept, the calibration slope, ECE, and Nagelkerke $\mathrm{R^2}$ are non-collapsible metrics, whereas the O:E ratio, logloss, Brier score, accuracy, F1-score, TPR, TNR, PPV, NPV, and net benefit are collapsible.

      \medskip
      \medskip
  
    The AUC is a non-collapsible metric because under mixtures of subpopulations it decomposes into within- and cross-group AUC components. Its overall value cannot always be recovered as a weighted average of within-group AUCs. This may lead to counter-intuitive scenarios such as each subgroup AUC being smaller than the overall AUC.

      \medskip
      \medskip

    AUC collapsibility holds when the distribution of estimated probabilities among individuals without the event are identical across subgroups. Non-collapsibility is therefore driven by distributional variations in estimated probabilities, arising from case-mix variation or miscalibration differences across subgroups.
  
  \medskip
  \medskip  

    Fairness evaluations based on differences between subgroups and overall performance, or across subgroups, are complex, as these quantities are not necessarily directly comparable, and differences may arise naturally.
    
      \medskip
      \medskip
       Transparent reporting is therefore essential. In the context of non-collapsible metrics, we underscore the importance of reporting population characteristics, outcome distributions, and fairness-related design choices. Adherence to TRIPOD+AI promotes transparency and reproducibility in fairness evaluations \cite{collins2024tripod+}.

\end{discussion_box}

\newpage
\section*{Acknowledgements}
JM is funded by a Clarendon Fund Scholarship at the University of Oxford. GSC and RDR are supported by the National Institute for Health and Care Research (NIHR) Biomedical Research Centre: Birmingham at University Hospitals Birmingham NHS Foundation Trust and the University of Birmingham. GSC and RDR are NIHR Senior Investigators. BVC is supported by Research Foundation – Flanders (FWO) grant G097322N, Kom Op Tegen Kanker grant 13583, and KU Leuven Internal Funds grant C24M/20/064. The funders of the study had no role in study design, data collection, data analysis, interpretation of data, writing of the report, or decision to submit the article for publication. The views expressed are those of the authors and not necessarily represent the views of the funding agencies.

\newpage
\bibliography{references}

@ARTICLE{Kallus2019-yl,
  title   = "The fairness of risk scores beyond classification: Bipartite
             ranking and the xauc metric",
  author  = "Kallus, Nathan and Zhou, Angela",
  journal = "Advances in neural information processing systems",
  volume  =  32,
  year    =  2019
}

@article{huitfeldt2019collapsibility,
  title={On the collapsibility of measures of effect in the counterfactual causal framework},
  author={Huitfeldt, Anders and Stensrud, Mats J and Suzuki, Etsuji},
  journal={Emerging themes in epidemiology},
  volume={16},
  number={1},
  pages={1},
  year={2019},
  publisher={Springer}
}

@ARTICLE{Faraggi2002-hm,
  title     = "Estimation of the area under the {ROC} curve",
  author    = "Faraggi, David and Reiser, Benjamin",
  journal   = "Stat. Med.",
  publisher = "Wiley",
  volume    =  21,
  number    =  20,
  pages     = "3093--3106",
  month     =  oct,
  year      =  2002,
  language  = "en"
}

@ARTICLE{Van-Calster2024-jw,
  title         = "Evaluation of performance measures in predictive artificial intelligence models to support medical decisions: overview and guidance",
  author        = "Van Calster, Ben and Collins, Gary S and Vickers, Andrew J and Wynants, Laure and Kerr, Kathleen F and Barre{\~n}ada, Lasai and Varoquaux, Gael and Singh, Karandeep and Moons, Karel GM and Hernandez-Boussard, Tina and others",
  journal       = "The Lancet Digital Health",
  publisher = "Elsevier",
  year          =  2025,
}

@article{matos2025critical,
  title={Critical Appraisal of Fairness Metrics in Clinical Predictive AI},
  author={Matos, Jo{\~a}o and Van Calster, Ben and Celi, Leo Anthony and Dhiman, Paula and Gichoya, Judy Wawira and Riley, Richard D and Russell, Chris and Khalid, Sara and Collins, Gary S},
  journal={The Lancet Digital Health},
  year={2026},
}

@article{simpson1951interpretation,
  title={The interpretation of interaction in contingency tables},
  author={Simpson, Edward H},
  journal={Journal of the Royal Statistical Society: Series B (Methodological)},
  volume={13},
  number={2},
  pages={238--241},
  year={1951},
  publisher={Wiley Online Library}
}

@article{blyth1972simpson,
  title={On Simpson's paradox and the sure-thing principle},
  author={Blyth, Colin R},
  journal={Journal of the American Statistical Association},
  volume={67},
  number={338},
  pages={364--366},
  year={1972},
  publisher={Taylor \& Francis}
}

@article{messick1981reversal,
  title={A reversal paradox.},
  author={Messick, David M and Van de Geer, John P},
  journal={Psychological Bulletin},
  volume={90},
  number={3},
  pages={582},
  year={1981},
  publisher={American Psychological Association}
}

@article{good1987amalgamation,
  title={The amalgamation and geometry of two-by-two contingency tables},
  author={Good, Irving John and Mittal, Yashaswini},
  journal={The Annals of Statistics},
  pages={694--711},
  year={1987},
  publisher={JSTOR}
}

@article{denny2017fallacy,
  title={The fallacy of the average: on the ubiquity, utility and continuing novelty of Jensen's inequality},
  author={Denny, Mark},
  journal={Journal of Experimental Biology},
  volume={220},
  number={2},
  pages={139--146},
  year={2017},
  publisher={The Company of Biologists Ltd}
}

@article{sudjianto2025decomposing,
  title={Decomposing Global AUC into Cluster-Level Contributions for Localized Model Diagnostics},
  author={Sudjianto, Agus and Liu, Alice J},
  journal={arXiv preprint arXiv:2508.07495},
  year={2025}
}

@article{nashef2012euroscore,
  title={Euroscore ii},
  author={Nashef, Samer AM and Roques, Fran{\c{c}}ois and Sharples, Linda D and Nilsson, Johan and Smith, Christopher and Goldstone, Antony R and Lockowandt, Ulf},
  journal={European journal of cardio-thoracic surgery},
  volume={41},
  number={4},
  pages={734--745},
  year={2012},
  publisher={Oxford University Press}
}

@article{gonccalves2014roc,
  title={ROC curve estimation: An overview},
  author={Gon{\c{c}}alves, Luzia and Subtil, Ana and Oliveira, M Ros{\'a}rio and de Zea Bermudez, Patricia},
  journal={REVSTAT-Statistical journal},
  volume={12},
  number={1},
  pages={1--20},
  year={2014}
}

@article{van2016calibration,
  title={A calibration hierarchy for risk models was defined: from utopia to empirical data},
  author={Van Calster, Ben and Nieboer, Daan and Vergouwe, Yvonne and De Cock, Bavo and Pencina, Michael J and Steyerberg, Ewout W},
  journal={Journal of clinical epidemiology},
  volume={74},
  pages={167--176},
  year={2016},
  publisher={Elsevier}
}

@article{tseng2021spectrum,
  title={Spectrum bias in algorithms derived by artificial intelligence: a case study in detecting aortic stenosis using electrocardiograms},
  author={Tseng, Andrew S and Shelly-Cohen, Michal and Attia, Itzhak Z and Noseworthy, Peter A and Friedman, Paul A and Oh, Jae K and Lopez-Jimenez, Francisco},
  journal={European Heart Journal-Digital Health},
  volume={2},
  number={4},
  pages={561--567},
  year={2021},
  publisher={Oxford University Press}
}

@article{janssens2020reflection,
  title={Reflection on modern methods: Revisiting the area under the ROC Curve},
  author={Janssens, A Cecile JW and Martens, Forike K},
  journal={International journal of epidemiology},
  volume={49},
  number={4},
  pages={1397--1403},
  year={2020},
  publisher={Oxford University Press}
}

@article{martens2016small,
  title={Small improvement in the area under the receiver operating characteristic curve indicated small changes in predicted risks},
  author={Martens, Forike K and Tonk, Elisa CM and Kers, Jannigje G and Janssens, A Cecile JW},
  journal={Journal of clinical epidemiology},
  volume={79},
  pages={159--164},
  year={2016},
  publisher={Elsevier}
}

@article{pencina2017discrimination,
  title={Discrimination slope and integrated discrimination improvement--properties, relationships and impact of calibration},
  author={Pencina, Michael J and Fine, Jason P and D'Agostino Sr, Ralph B},
  journal={Statistics in Medicine},
  volume={36},
  number={28},
  pages={4482--4490},
  year={2017},
  publisher={Wiley Online Library}
}

@article{van2016new,
  title={A new concordance measure for risk prediction models in external validation settings},
  author={van Klaveren, David and G{\"o}nen, Mithat and Steyerberg, Ewout W and Vergouwe, Yvonne},
  journal={Statistics in medicine},
  volume={35},
  number={23},
  pages={4136--4152},
  year={2016},
  publisher={Wiley Online Library}
}

@article{sadatsafavi2022model,
  title={Model-based ROC curve: examining the effect of case mix and model calibration on the ROC plot},
  author={Sadatsafavi, Mohsen and Saha-Chaudhuri, Paramita and Petkau, John},
  journal={Medical Decision Making},
  volume={42},
  number={4},
  pages={487--499},
  year={2022},
  publisher={SAGE Publications Sage CA: Los Angeles, CA}
}

@article{vergouwe2010external,
  title={External validity of risk models: use of benchmark values to disentangle a case-mix effect from incorrect coefficients},
  author={Vergouwe, Yvonne and Moons, Karel GM and Steyerberg, Ewout W},
  journal={American journal of epidemiology},
  volume={172},
  number={8},
  pages={971--980},
  year={2010},
  publisher={Oxford University Press}
}

@article{collins2024tripod+,
  title={TRIPOD+ AI statement: updated guidance for reporting clinical prediction models that use regression or machine learning methods},
  author={Collins, Gary S and Moons, Karel GM and Dhiman, Paula and Riley, Richard D and Beam, Andrew L and Van Calster, Ben and Ghassemi, Marzyeh and Liu, Xiaoxuan and Reitsma, Johannes B and Van Smeden, Maarten and others},
  journal={bmj},
  volume={385:e078378},
  year={2024},
  publisher={British Medical Journal Publishing Group}
}

@article{kleinberg2016inherent,
  title={Inherent trade-offs in the fair determination of risk scores},
  author={Kleinberg, Jon and Mullainathan, Sendhil and Raghavan, Manish},
  journal={arXiv preprint arXiv:1609.05807},
  year={2016}
}

@article{hernan2011simpson,
  title={The Simpson's paradox unraveled},
  author={Hern{\'a}n, Miguel A and Clayton, David and Keiding, Niels},
  journal={International journal of epidemiology},
  volume={40},
  number={3},
  pages={780--785},
  year={2011},
  publisher={Oxford University Press}
}

@article{van2026navigating,
  title={Navigating fairness in artificial intelligence-based prediction models: theoretical constructs and practical applications},
  author={van der Meijden, Siri L and Wang, Yuqing and Ng, Madelena Y and Arbous, Sesmu M and Geerts, Bart F and Steyerberg, Ewout W and Hernandez-Boussard, Tina},
  journal={The Lancet Digital Health},
  year={2026},
  publisher={Elsevier}
}

@book{zhou2009statistical,
  title={Statistical Methods in Diagnostic Medicine},
  author={Zhou, X.H. and McClish, D.K. and Obuchowski, N.A.},
  isbn={9780470317921},
  lccn={2002514697},
  series={Wiley Series in Probability and Statistics},
  url={https://books.google.com/books?id=ijN_Dlx7wmoC},
  year={2009},
  publisher={Wiley}
}

@article{austin2012interpreting,
  title={Interpreting the concordance statistic of a logistic regression model: relation to the variance and odds ratio of a continuous explanatory variable},
  author={Austin, Peter C and Steyerberg, Ewout W},
  journal={BMC medical research methodology},
  volume={12},
  number={1},
  pages={82},
  year={2012},
  publisher={Springer}
}

@article{de2023propensity,
  title={Propensity-based standardization to enhance the validation and interpretation of prediction model discrimination for a target population},
  author={de Jong, Valentijn MT and Hoogland, Jeroen and Moons, Karel GM and Riley, Richard D and Nguyen, Tri-Long and Debray, Thomas PA},
  journal={Statistics in Medicine},
  volume={42},
  number={19},
  pages={3508--3528},
  year={2023},
  publisher={Wiley Online Library}
}

@article{riley2025decomposition1,
  title={A decomposition of Fisher’s information to inform sample size for developing or updating fair and precise clinical prediction models for individual risk—part 1: binary outcomes},
  author={Riley, Richard D and Collins, Gary S and Whittle, Rebecca and Archer, Lucinda and Snell, Kym IE and Dhiman, Paula and Kirton, Laura and Legha, Amardeep and Liu, Xiaoxuan and Denniston, Alastair K and others},
  journal={Diagnostic and prognostic research},
  volume={9},
  number={1},
  pages={14},
  year={2025},
  publisher={Springer}
}

@article{riley2025decomposition2,
  title={A decomposition of Fisher’s information to inform sample size for developing or updating fair and precise clinical prediction models—part 2: time-to-event outcomes},
  author={Riley, Richard D and Collins, Gary S and Archer, Lucinda and Whittle, Rebecca and Legha, Amardeep and Kirton, Laura and Dhiman, Paula and Sadatsafavi, Mohsen and Adderley, Nicola J and Alderman, Joseph and others},
  journal={Diagnostic and Prognostic Research},
  volume={9},
  number={1},
  pages={33},
  year={2025},
  publisher={Springer}
}

@article{whittle2026decomposition3,
  title={A decomposition of Fisher’s information to inform sample size for developing or updating fair and precise clinical prediction models--part 3: continuous outcomes},
  author={Whittle, Rebecca and Riley, Richard D and Archer, Lucinda and Collins, Gary S and Legha, Amardeep and Snell, Kym IE and Ensor, Joie},
  journal={Diagnostic and Prognostic Research},
  volume={10},
  number={1},
  pages={11},
  year={2026},
  publisher={Springer}
}

\newpage
\addtocontents{toc}{\protect\setcounter{tocdepth}{1}}
\appendix
\section{Appendix}
\subsection{Collapsibility Proofs}
\label{app_collap}
We investigate the performance measures summarised in Table~\ref{tab:perf_measures}.

\subsubsection{O:E Ratio}
The observed-to-expected ratio (O:E) compares the number of observed outcomes to the 
expected number of outcomes under the model.

\begin{align*}
\mathrm{O:E}
&\triangleq \frac{O}{E},
\qquad O = \sum_{i=1}^N y_i, \quad E = \sum_{i=1}^N \hat p_i, \\
\mathrm{with}\quad
\mathrm{(O:E)}_g
&\triangleq \frac{O_g}{E_g}, \quad O_g = \sum_{i\in g} y_i, \quad E_g = \sum_{i\in g} \hat p_i.
\end{align*}

Substituting group-wise counts:

\begin{align*}
\mathrm{O:E}
&= \frac{O_A + O_B}{E_A + E_B} \\
&= \frac{1}{E_A + E_B}
\Big[
\underbrace{O_A}_{\,E_A\,\mathrm{(O:E)}_A}
+
\underbrace{O_B}_{\,E_B\,\mathrm{(O:E)}_B}
\Big] \\
&= \frac{E_A}{E_A+E_B}\,\mathrm{(O:E)}_A
 + \frac{E_B}{E_A+E_B}\,\mathrm{(O:E)}_B \;\;\; \Box
\end{align*}

Thus the O:E ratio satisfies~\ref{eq_collap1} with weights 
\[
w_g = \frac{E_g}{\sum_h E_h}.
\]

Therefore, O:E is collapsible with weights proportional to the group-expected counts under the model.

\subsubsection{Calibration Intercept}
\newcommand{\logit}{\operatorname{logit}}
\newcommand{\expit}{\operatorname{expit}}

Calibration intercept $\alpha$ is usually defined in terms of:

\[
\mathrm{logit}(y) = \alpha + 1 \times \mathrm{logit}(\hat{p}).
\]

Let groups $g\in\{A,B\}$ have sizes $n_g$ and weights $w_g=n_g/n$ with $n=n_A+n_B$.
For observations $(y_i^g,\hat p_i^g)$ with $0<\hat p_i^g<1$, define
\[
  \expit(t) = \frac{1}{1+e^{-t}},\qquad
  \logit(p) = \log\Big(\frac{p}{1-p}\Big), \qquad 
  u_i^g(\alpha) = \expit\!\big(\alpha+\logit(\hat p_i^g)\big),
\]
\[
  f_g(\alpha)=\frac{1}{n_g}\sum_{i=1}^{n_g}u_i^g(\alpha).
\]

The calibration intercept $\alpha_g$ solves
\[
  f_g(\alpha_g)=\bar y_g,\qquad \bar y_g=\frac{1}{n_g}\sum_{i=1}^{n_g}y_i^g.
\]
For the pooled sample one can define
\[
  f(\alpha)=\frac{1}{n}\sum_{g\in\{A,B\}}\sum_{i=1}^{n_g}u_i^g(\alpha)
            = w_A f_A(\alpha)+w_B f_B(\alpha),
\]
and 
\[
\bar y=\frac{1}{n}\sum_{g,i}y_i^g=w_A\bar y_A+w_B\bar y_B.
\]

\paragraph{Lemma 1 (estimating equation).}
The pooled calibration intercept $\alpha_{\mathrm{all}}$ is the unique solution of
\[
  f(\alpha)=\bar y.
\]

\emph{Proof.}
The \emph{log-likelihood} is
\[
  \ell(\alpha)
  =\sum_{g,i}\Big\{y_i^g\log u_i^g(\alpha)+(1-y_i^g)\log\big(1-u_i^g(\alpha)\big)\Big\},
\]
and the \emph{score}, the derivative of the log-likelihood with respect to $\alpha$, is
\[
  S(\alpha)=\ell'(\alpha)
  =\sum_{g,i}\big(y_i^g-u_i^g(\alpha)\big),
\]
since $\frac{\partial u_i^g(\alpha)}{\partial \alpha}
= u_i^g(\alpha)\bigl(1-u_i^g(\alpha)\bigr)$.
The maximum likelihood estimator satisfies the score equation $S(\alpha)=0$, that is
\[
  \frac{1}{n}\sum_{g,i}u_i^g(\alpha)
  = \frac{1}{n}\sum_{g,i}y_i^g
  \quad\Longleftrightarrow\quad
  f(\alpha)=\bar y.
\]

Uniqueness follows because $\ell(\alpha)$ is strictly concave in $\alpha$ and admits a unique maximiser, equivalently a unique root of $S(\alpha)=0$. \qed

\paragraph{Lemma 2 (monotonicity).}
Each $f_g$ is continuous and strictly increasing; hence $f$ is continuous and strictly increasing.

\emph{Proof.}
$f_g'(\alpha)=\frac{1}{n_g}\sum_{i} u_i^g(\alpha)\{1-u_i^g(\alpha)\}>0$. Linearity gives the result for $f$. \qed

\paragraph{Proposition (convex bounds).}
Let $\alpha_-=\min\{\alpha_A,\alpha_B\}$ and $\alpha_+=\max\{\alpha_A,\alpha_B\}$.
Then the overall intercept satisfies
\[
  \alpha_{\mathrm{all}}\in[\alpha_-,\alpha_+].
\]

\emph{Proof.}
Assume, without loss of generality, that \ $\alpha_A\le \alpha_B$. Then
\[
  f(\alpha_A)=w_A f_A(\alpha_A)+w_B f_B(\alpha_A)
            \le w_A \bar y_A + w_B \bar y_B = \bar y,
\]
since $f_B$ is increasing and $\alpha_A\le \alpha_B \Longrightarrow f_B(\alpha_A) \leq f_B(\alpha_B)$. Similarly,
\[
  f(\alpha_B)=w_A f_A(\alpha_B)+w_B f_B(\alpha_B)
            \ge w_A \bar y_A + w_B \bar y_B = \bar y,
\]
since $f_A$ is increasing and $\alpha_B\ge \alpha_A$. By Lemma 2 and the intermediate value theorem, there is a unique \(\alpha_{\mathrm{all}}\in[\alpha_A,\alpha_B]\) with $f(\alpha_{\mathrm{all}})=\bar y$. Hence, \(f(\alpha_A)\;\le\;\bar y\;\le\;f(\alpha_B)\), and for every $\alpha_A < \alpha_{\mathrm{all}} < \alpha_B$ we have \(f(\alpha_A) \;\leq\; f(\alpha_{\mathrm{all}}) \;\leq\; f(\alpha_B).\) \qed

\medskip
\noindent\textbf{Remark.}
The only property established is the bounding relation $\alpha_{\mathrm{all}}\in[\alpha_A,\alpha_B]$. 
With this, we are not claiming that the calibration intercept is \emph{necessarily collapsible}: in general,
$\alpha_{\mathrm{all}}$ is not expected to be expressed as a convex or linear combination of  $\alpha_A$ and $\alpha_B$.

\subsubsection{Calibration Slope}
The calibration slope $\beta$ is estimated from the regression
\[
\mathrm{logit}(y_i) = \alpha + \beta \,\mathrm{logit}(\hat p_i) + \varepsilon_i,
\]
with ideal calibration corresponding to \(\beta=1\).  

The calibration slope $\beta$ is estimated as a regression coefficient from a logistic regression model fitted to the entire dataset. Logistic regression coefficients are generally non-collapsible, meaning that the coefficient estimated from the pooled data is not, in general, a weighted average of coefficients estimated within subgroups. Consequently, the calibration slope is also expected to be non-collapsible.

\subsubsection{Expected Calibration Error (ECE)}
The expected calibration error measures the average absolute deviation between predicted and observed outcome rates over bins of estimated probability.

\begin{align*}
\mathrm{ECE}
&\triangleq \sum_{b=1}^B \frac{n_b}{N}\,
\Big|\,\frac{1}{n_b}\sum_{i\in b} y_i - \frac{1}{n_b}\sum_{i\in b}\hat p_i \,\Big|, \\
\mathrm{with}\quad
\mathrm{ECE}_g
&\triangleq \sum_{b=1}^B \frac{n_{gb}}{N_g}\,
\Big|\,\frac{1}{n_{gb}}\sum_{i\in (g,b)} y_i - \frac{1}{n_{gb}}\sum_{i\in (g,b)}\hat p_i \,\Big|.
\end{align*}

Although ECE is an average across bins, the absolute value inside the bin definition 
prevents linear aggregation across groups:
\[
\mathrm{ECE} \neq \sum_g \frac{N_g}{N}\,\mathrm{ECE}_g \quad \text{in general}.
\]

Thus ECE is expected to be \emph{non-collapsible}.

\subsubsection{Logloss / Cross-Entropy}
Logloss (cross-entropy) is the empirical average of the per-instance negative log-likelihood; it is collapsible.

\begin{align*}
\mathrm{LogLoss}
&\triangleq \frac{1}{N}\sum_{i=1}^{N}\Big[-y_i\log \hat p_i - (1-y_i)\log(1-\hat p_i)\Big], \\
\mathrm{and}\quad
\mathrm{LogLoss}_g
&\triangleq \frac{1}{N_g}\sum_{i\in g}\Big[-y_i\log \hat p_i - (1-y_i)\log(1-\hat p_i)\Big].
\end{align*}

Partitioning by groups:
\begin{align*}
\mathrm{LogLoss}
&= \frac{1}{N}\sum_{g}\sum_{i\in g}\Big[-y_i\log \hat p_i - (1-y_i)\log(1-\hat p_i)\Big] \\
&= \sum_{g}\frac{N_g}{N}\;\underbrace{\frac{1}{N_g}\sum_{i\in g}\Big[-y_i\log \hat p_i - (1-y_i)\log(1-\hat p_i)\Big]}_{\,\mathrm{LogLoss}_g\,}
= \sum_{g} p^g\,\mathrm{LogLoss}_g \;\;\; \Box
\end{align*}

Thus Logloss satisfies~\ref{eq_collap1} with weights \(w_g=p^g=N_g/N\).

\subsubsection{Brier Score}

The Brier score is the average squared error; it is collapsible.

\begin{align*}
\mathrm{Brier}
&\triangleq \frac{1}{N}\sum_{i=1}^{N}\big(\hat p_i - y_i\big)^2 \quad
\mathrm{and}\quad \mathrm{Brier}_g
\triangleq \frac{1}{N_g}\sum_{i\in g}\big(\hat p_i - y_i\big)^2.
\end{align*}

Partitioning by groups:
\begin{align*}
\mathrm{Brier}
&= \frac{1}{N}\sum_{g}\sum_{i\in g}\big(\hat p_i - y_i\big)^2
= \sum_{g}\frac{N_g}{N}\;\underbrace{\frac{1}{N_g}\sum_{i\in g}\big(\hat p_i - y_i\big)^2}_{\,\mathrm{Brier}_g\,}
= \sum_{g} p^g\,\mathrm{Brier}_g \;\;\; \Box
\end{align*}

Thus the Brier score satisfies~\ref{eq_collap1} with weights \(w_g=p^g\).

\subsubsection{Nagelkerke \texorpdfstring{$R^2$}{R2}}

Nagelkerke's \(R^2\) (for binary outcomes) rescales Cox–Snell’s \(R^2\) to \([0,1]\):
\[
R^2_{\mathrm{N}} 
= \frac{1 - \exp\!\Big(\tfrac{2}{N}\,(\ell_0 - \ell_1)\Big)}
{1 - \exp\!\Big(\tfrac{2}{N}\,\ell_0\Big)},
\quad
\ell_1=\sum_{i=1}^N \log p_{\theta}(y_i\mid x_i),\quad
\ell_0=\sum_{i=1}^N \log p_{\theta_0}(y_i).
\].

Here \(\ell_1\) denotes the log-likelihood of the fitted model with parameter 
\(\theta\), and \(\ell_0\) the log-likelihood of the null model (intercept-only, 
with parameter \(\theta_0\)). The term \(p_{\theta}(y_i\mid x_i)\) is the model-estimated probability of observing outcome \(y_i\) given covariates \(x_i\) under parameter \(\theta\). Group-wise versions are then:
\[
R^2_{\mathrm{N},g} 
= \frac{1 - \exp\!\Big(\tfrac{2}{N_g}\,(\ell_{0g} - \ell_{1g})\Big)}
{1 - \exp\!\Big(\tfrac{2}{N_g}\,\ell_{0g}\Big)},
\quad
\ell_{1}=\sum_g \ell_{1g},\;\; \ell_{0}=\sum_g \ell_{0g},\;\; N=\sum_g N_g.
\]

In general,
\[
R^2_{\mathrm{N}} 
= 
\frac{1 - \exp\!\Big(\tfrac{2}{N}\sum_g(\ell_{0g}-\ell_{1g})\Big)}
{1 - \exp\!\Big(\tfrac{2}{N}\sum_g \ell_{0g}\Big)}
\;\;\neq\;\;
\sum_g w_g\,
\frac{1 - \exp\!\Big(\tfrac{2}{N_g}(\ell_{0g}-\ell_{1g})\Big)}
{1 - \exp\!\Big(\tfrac{2}{N_g}\ell_{0g}\Big)}.
\]

Because the mapping \((\ell_{0},\ell_{1},N)\mapsto R^2_{\mathrm{N}}\) applies nonlinear exponentials with different normalisations (\(1/N\) vs \(1/N_g\)) to sums of log-likelihoods, the overall index is not a convex combination of group indices except under potentially restrictive special cases. Therefore, Nagelkerke \(R^2\) does not satisfy ~\ref{eq_collap1} in general and is expected to be non-collapsible. If an amalgamation paradox is observed to occur, the measure is proven to be non-collapsible.

\subsubsection{Classification accuracy}

Classification accuracy is collapsible with weights given by the group sizes.

\begin{align*}
\mathrm{Acc}
&\triangleq \frac{TP + TN}{N},
\qquad N = TP + TN + FP + FN, \\
\mathrm{and}\quad
\mathrm{Acc}_g &\triangleq \frac{TP_g + TN_g}{N_g}.
\end{align*}

Substituting group-wise counts:

\begin{align*}
\mathrm{Acc}
&= \frac{(TP_A + TN_A) + (TP_B + TN_B)}{N_A + N_B} \\
&= \frac{1}{N_A + N_B}
\Big[
\underbrace{TP_A + TN_A}_{\,N_A\,\mathrm{Acc}_A}
+
\underbrace{TP_B + TN_B}_{\,N_B\,\mathrm{Acc}_B}
\Big] \\
&= \frac{N_A}{N_A + N_B}\,\mathrm{Acc}_A
 + \frac{N_B}{N_A + N_B}\,\mathrm{Acc}_B \;\;\; \Box
\end{align*}

Thus accuracy satisfies~\ref{eq_collap1} with weights \(w_g = N_g/N = p^g\).

\subsubsection{F1-score}
F1-score is the harmonic mean of sensitivity and precision, which can also be defined as:

\begin{align*}
\mathrm{F1}
&\triangleq \frac{2TP}{2TP + FP + FN} 
\quad \mathrm{and} \quad
\mathrm{F1}_g \triangleq \frac{2TP_g}{\,2TP_g + FP_g + FN_g\,}.
\end{align*}

Substituting group-wise counts and writing the overall F1 as a ratio of linear sums:

\begin{align*}
\mathrm{F1}
&= \frac{\,2TP_A + 2TP_B\,}{\,(2TP_A + FP_A + FN_A) + (2TP_B + FP_B + FN_B)\,} \\[6pt]
&= 
\underbrace{\Big(\frac{\,2TP_A\,}{\,2TP_A + FP_A + FN_A\,}\Big)}_{\,F1_A\,}\cdot
\frac{\,2TP_A + FP_A + FN_A\,}{\sum_{g}(2TP_g + FP_g + FN_g)}
\\[6pt]
&\quad +
\underbrace{\Big(\frac{\,2TP_B\,}{\,2TP_B + FP_B + FN_B\,}\Big)}_{\,F1_B\,}\cdot
\frac{\,2TP_B + FP_B + FN_B\,}{\sum_{g}(2TP_g + FP_g + FN_g)}
\\[6pt]
&= \sum_{g\in\{A,B\}}
\underbrace{\frac{\,2TP_g + FP_g + FN_g\,}{\sum_{h}(2TP_h + FP_h + FN_h)}}_{\,w_g\,}
\;\mathrm{F1}_g \;\;\; \Box
\end{align*}

Thus F1 satisfies~\ref{eq_collap1} with weights
\(w_g = \dfrac{2TP_g + FP_g + FN_g}{\sum_h (2TP_h + FP_h + FN_h)}\).

\subsubsection{Sensitivity / Recall / TPR}

Sensitivity (TPR) is collapsible with weights proportional to the group-specific prevalence of the outcome.

\begin{align*}
\mathrm{TPR} 
&\triangleq \frac{TP}{N^{+}}, 
\qquad N^{+} = TP + FN, \\
\mathrm{and}\quad
\mathrm{TPR}_g &\triangleq \frac{TP_g}{N^{+}_g},
\quad N^{+}_g = TP_g + FN_g.
\end{align*}

Substituting group-wise counts:

\begin{align*}
\mathrm{TPR}
&= \frac{TP_A + TP_B}{N^{+}_A + N^{+}_B} \\
&= \frac{1}{N^{+}_A + N^{+}_B}
\Big[
\underbrace{TP_A}_{\,N^{+}_A\,\mathrm{TPR}_A}
+
\underbrace{TP_B}_{\,N^{+}_B\,\mathrm{TPR}_B}
\Big] \\
&= \frac{N^{+}_A}{N^{+}}\,\mathrm{TPR}_A
 + \frac{N^{+}_B}{N^{+}}\,\mathrm{TPR}_B \;\;\; \Box
\end{align*}

Thus TPR satisfies~\ref{eq_collap1} with weights 
\[
w_g = \frac{N^{+}_g}{N^{+}} = \frac{p^g \pi_g}{\pi},
\]
where \(p^g = N_g/N\), \(\pi_g = N^{+}_g/N_g\) and \(\pi = N^{+}/N\).

\subsubsection{Specificity / TNR}

Specificity (TNR) is collapsible with weights proportional to the group-specific rate of actual negatives.

\begin{align*}
\mathrm{TNR} 
&\triangleq \frac{TN}{N^{-}}, 
\qquad N^{-} = TN + FP, \\
\mathrm{and}\quad
\mathrm{TNR}_g &\triangleq \frac{TN_g}{N^{-}_g},
\quad N^{-}_g = TN_g + FP_g.
\end{align*}

Substituting group-wise counts:

\begin{align*}
\mathrm{TNR}
&= \frac{TN_A + TN_B}{N^{-}_A + N^{-}_B} \\
&= \frac{1}{N^{-}_A + N^{-}_B}
\Big[
\underbrace{TN_A}_{\,N^{-}_A\,\mathrm{TNR}_A}
+
\underbrace{TN_B}_{\,N^{-}_B\,\mathrm{TNR}_B}
\Big] \\
&= \frac{N^{-}_A}{N^{-}}\,\mathrm{TNR}_A
 + \frac{N^{-}_B}{N^{-}}\,\mathrm{TNR}_B \;\;\; \Box
\end{align*}

Thus TNR satisfies~\ref{eq_collap1} with weights 
\[
w_g = \frac{N^{-}_g}{N^{-}} = \frac{p^g (1-\pi_g)}{1-\pi}.
\]

\subsubsection{PPV / Precision}

PPV is collapsible with weights proportional to the group-specific rate of predicted positives.

\begin{align*}
\mathrm{PPV} 
&\triangleq \frac{TP}{\hat N^{+}}, 
\qquad \hat N^{+} = TP + FP, \\
\mathrm{and}\quad
\mathrm{PPV}_g &\triangleq \frac{TP_g}{\hat N^{+}_g},
\quad \hat N^{+}_g = TP_g + FP_g.
\end{align*}

Substituting group-wise counts:

\begin{align*}
\mathrm{PPV}
&= \frac{TP_A + TP_B}{\hat N^{+}_A + \hat N^{+}_B} \\
&= \frac{1}{\hat N^{+}_A + \hat N^{+}_B}
\Big[
\underbrace{TP_A}_{\,\hat N^{+}_A\,\mathrm{PPV}_A}
+
\underbrace{TP_B}_{\,\hat N^{+}_B\,\mathrm{PPV}_B}
\Big] \\
&= \frac{\hat N^{+}_A}{\hat N^{+}}\,\mathrm{PPV}_A
 + \frac{\hat N^{+}_B}{\hat N^{+}}\,\mathrm{PPV}_B \;\;\; \Box
\end{align*}

Thus PPV satisfies~\ref{eq_collap1} with weights 
\[
w_g = \frac{\hat N^{+}_g}{\hat N^{+}} = \frac{p^g \hat\pi_g}{\hat\pi},
\]
where \(\hat\pi_g = \hat N^{+}_g/N_g\) is the group-specific predicted positive rate, and \(\hat\pi = \hat N^{+}/N\).

\subsubsection{NPV}
NPV is collapsible with weights proportional to the group-specific rate of predicted negatives.

\begin{align*}
\mathrm{NPV} 
&\triangleq \frac{TN}{\hat N^{-}}, 
\qquad \hat N^{-} = TN + FN, \\
\mathrm{and}\quad
\mathrm{NPV}_g &\triangleq \frac{TN_g}{\hat N^{-}_g},
\quad \hat N^{-}_g = TN_g + FN_g.
\end{align*}

Substituting group-wise counts:

\begin{align*}
\mathrm{NPV}
&= \frac{TN_A + TN_B}{\hat N^{-}_A + \hat N^{-}_B} \\
&= \frac{1}{\hat N^{-}_A + \hat N^{-}_B}
\Big[
\underbrace{TN_A}_{\,\hat N^{-}_A\,\mathrm{NPV}_A}
+
\underbrace{TN_B}_{\,\hat N^{-}_B\,\mathrm{NPV}_B}
\Big] \\
&= \frac{\hat N^{-}_A}{\hat N^{-}}\,\mathrm{NPV}_A
 + \frac{\hat N^{-}_B}{\hat N^{-}}\,\mathrm{NPV}_B \;\;\; \Box
\end{align*}

Thus NPV satisfies~\ref{eq_collap1} with weights 
\[
w_g = \frac{\hat N^{-}_g}{\hat N^{-}} = \frac{p^g (1-\hat\pi_g)}{1-\hat\pi},
\]
where \(\hat\pi_g = \hat N^{+}_g/N_g\) and \(\hat\pi = \hat N^{+}/N\).

\subsubsection{Net Benefit}

Net Benefit ($NB$) is often used in decision curve analysis as a summary of clinical usefulness. It combines true positives and false positives into a single measure, weighted by the relative harms of false positives at a given threshold, $t$. Here we show that $NB$ is a collapsible measure.

\begin{align*}
\mathrm{NB} 
&\triangleq \frac{TP}{N} - c\,\frac{FP}{N}, 
\qquad c = t(1-t), \\
\mathrm{and}\quad
\mathrm{NB}_g &\triangleq \frac{TP_g}{N_g} - c\,\frac{FP_g}{N_g}.
\end{align*}

Substituting group-wise counts into the definition:

\begin{align*}
\mathrm{NB} 
&= \frac{TP_A + TP_B}{N} - c\,\frac{FP_A + FP_B}{N} \\[6pt]
&= \frac{1}{N}\Big[(TP_A - c\,FP_A) + (TP_B - c\,FP_B)\Big] \\[6pt]
&= \frac{1}{N}\Big[
\underbrace{TP_A - c\,FP_A}_{\,N_A\,\mathrm{NB}_A}
+
\underbrace{TP_B - c\,FP_B}_{\,N_B\,\mathrm{NB}_B}
\Big] \\[6pt]
&= \frac{N_A}{N}\,\mathrm{NB}_A + \frac{N_B}{N}\,\mathrm{NB}_B \\[6pt]
&= p^A\,\mathrm{NB}_A + p^B\,\mathrm{NB}_B
= \sum_{g\in\{A,B\}} p^g\,\mathrm{NB}_g \;\;\; \Box
\end{align*}

Thus NB satisfies~\ref{eq_collap1} with weights \(w_g = p^g\), proving that NB is collapsible.

\subsection{Non-collapsibilty Counterexamples}

\subsubsection{Calibration Slope}
We here present a counterexample showing that the calibration slope is not a collapsible measure with respect to a grouping variable. We use two subpopulations of equal size ($n = 1{,}000$ in each subgroup). For subgroup $A$ a single predictor $X_A$ is drawn from a standard normal distribution,
\[
X_A \sim \mathcal N(0,1),
\]
and the event probability is defined by a logistic model,
\[
P(Y_A = 1 \mid X_A = x) = \frac{1}{1 + \exp(-2x)}.
\]
Binary outcomes $Y_A$ are then sampled from this Bernoulli model. For subgroup $B$ the same logistic relationship is used but the predictor distribution is shifted,
\[
X_B \sim \mathcal N(2,1),
\qquad
P(Y_B = 1 \mid X_B = x) = \frac{1}{1 + \exp(-2x)},
\]
and binary outcomes $Y_B$ are drawn accordingly.

Pooling the two groups changes the distribution of the linear predictor and the outcome in a nonlinear way, producing a overall calibration slope that is not a weighted average of the subgroup-specific slopes.

\renewcommand{\arraystretch}{1.3}
\begin{table}[H]
  \centering
  \caption{Toy example, calibration slope computed separately by subgroup and for the pooled sample.}
  \label{tab:cal-slope-example}
  \medskip
  \begin{tabular}{lcc}
    \hline
    \textbf{Subgroup} & $n$ & \textbf{Calibration slope} \\
    \hline
    A & 1{,}000 & 0.934 \\
    B & 1{,}000 & 0.963 \\
    Population & 2{,}000 & 0.979 \\
    \hline
  \end{tabular}
\end{table}

If the calibration slope were collapsible then the overall slope would equal a simple weighted average of the subgroup slopes, where the weights depend only on quantities such as sample size or outcome prevalence. Table~\ref{tab:cal-slope-example} shows that this is not the case: the overall slope (0.979) exceeds both subgroup values (0.934 and 0.963), so it cannot be written as a linear combination of them.

\qed

\subsubsection{Nagelkerke \texorpdfstring{$R^2$}{R2}}

We now show by counterexample that Nagelkerke's $R^2$ is not a collapsible measure. Data are generated similarly as in the calibration slope example. When the two groups are pooled, this heterogeneity in $X$ affects the null model likelihood and the fitted model likelihood in a nonlinear way, which leads to a overall Nagelkerke $R^2$ that is not a weighted average of the subgroup $R^2$ values.

\renewcommand{\arraystretch}{1.3} 
\begin{table}[H]
  \centering
  \caption{Toy example, Nagelkerke $R^2$ computed separately by subgroup and for the pooled sample.}
  \label{tab:nagelkerke-example}
  \medskip
  \begin{tabular}{lcc}
    \hline
    \textbf{Subgroup} & $n$ & $\mathrm{R^2}$ \\
    \hline
    A & 1,000 & 0.481 \\
    B & 1,000 & 0.440 \\
    Population & 2,000 & 0.623 \\
    \hline
  \end{tabular}
\end{table}

If Nagelkerke $R^2$ were collapsible with respect to the grouping variable then the overall $R^2$ would be a simple weighted average of the subgroup $R^2$ values (weights depending only on sample sizes or outcome prevalences). Table \ref{tab:nagelkerke-example} shows this is not true for the toy data: the overall $R^2$ (0.623) is larger than both subgroup values (0.481 and 0.440).

\qed

\subsubsection{Expected Calibration Error (ECE)}
We present a toy example showing how calibration assessed by the ECE can differ between subgroups and the population. We use two subpopulations of equal size ($n = 1{,}000$ in each subgroup). For subgroup $A$ a single predictor $X_A$ is drawn from a normal distribution,
\[
X_A \sim \mathcal{N}(0,1),
\]
and the event probability is defined by a logistic model,
\[
P(Y_A = 1 \mid X_A = x) = \frac{1}{1 + \exp\big(-\,( 0.7\,x )\big)}.
\]
Binary outcomes $Y_A$ are then sampled from this Bernoulli model. For subgroup $B$ the same logistic relationship is used but the predictor distribution is shifted,
\[
X_B \sim \mathcal{N}(2,1),
\qquad
P(Y_B = 1 \mid X_B = x) = \frac{1}{1 + \exp\big(-\,( 0.7\,x )\big)},
\]
and binary outcomes $Y_B$ are drawn accordingly. Estimated probabilities $\hat{p}$ are given by the model logistic probabilities and ECE is computed using quantile (equal-sized) bins with $10$ bins.

Pooling the two groups changes the empirical relationship between estimated probabilities and observed outcomes in a nonlinear way.

\renewcommand{\arraystretch}{1.3}
\begin{table}[H]
  \centering
  \caption{Toy example, ECE computed separately by subgroup and for the pooled sample.}
  \label{tab:ece-example}
  \medskip
  \begin{tabular}{lcc}
    \hline
    \textbf{Subgroup} & $n$ & ECE \\
    \hline
    A & 1{,}000 & 0.0378 \\
    B & 1{,}000 & 0.0379 \\
    Population & 2{,}000 & 0.0297 \\
    \hline
  \end{tabular}
\end{table}

If the ECE were collapsible then the overall ECE would equal a simple weighted average of the subgroup ECEs with weights depending only on quantities such as sample size or outcome prevalence. Table~\ref{tab:ece-example} shows that this is not the case: the overall ECE (0.0297) differs from the subgroup values (0.0378 and 0.0379), so it cannot be written as a simple linear combination of them.

\qed

\end{document}